\documentclass[11pt]{article}

\usepackage[final]{acl}

\usepackage{times}
\usepackage{latexsym}

\usepackage[T1]{fontenc}

\usepackage[utf8]{inputenc}

\usepackage{microtype}

\usepackage{inconsolata}

\usepackage{graphicx}

\usepackage[utf8]{inputenc}   
\usepackage{booktabs}         
\usepackage{amsfonts}         
\usepackage{amssymb}
\usepackage{nicefrac}         
\usepackage{microtype}        
\usepackage{xcolor}           
\usepackage{subcaption}
\usepackage{array}
\usepackage{multirow}
\usepackage{pifont}           
\usepackage{adjustbox}
\usepackage{makecell}
\usepackage[table]{xcolor}
\usepackage{placeins}
\usepackage{tcolorbox}
\usepackage{tabularx}
\usepackage{enumitem}
\usepackage{bm}
\usepackage{amsmath}
\usepackage{enumitem}

\usepackage{tikz}
\usetikzlibrary{arrows.meta,positioning,fit,calc,shapes.geometric,shapes.symbols}
\usepackage{subcaption}
\usepackage{fontawesome5}

\usepackage{hyperref}
\usepackage{xurl}

\definecolor{gold}{RGB}{218,165,32}
\definecolor{silver}{RGB}{0,0,139}
\definecolor{bronze}{RGB}{205,127,50}
\definecolor{professionalred}{RGB}{179,40,40}
\definecolor{lightblue}{rgb}{0.9,0.95,1.0}
\definecolor{lightgreen}{rgb}{0.92,0.98,0.92}
\definecolor{softpurple}{rgb}{0.95,0.92,1.0}

\newcommand{\secondplace}[1]{\textcolor{silver}{#1}}

\title{Chat-TS: Enhancing Multi-Modal Reasoning Over Time-Series and Natural Language Data}

\author{
  Paul Quinlan$^{1,3}$ \and Qingguo Li$^{2,3}$ \and Xiaodan Zhu$^{1,3}$ \\
  $^{1}$Electrical and Computer Engineering, Queen's University \\
  $^{2}$Mechanical and Materials Engineering, Queen's University \\
  $^{3}$Ingenuity Labs Research Institute, Queen's University \\
  \texttt{15pwq@queensu.ca, ql3@queensu.ca, xiaodan.zhu@queensu.ca}
}

\begin{document}
\maketitle

\begin{abstract}
Large language models are being rapidly deployed across many fields such as healthcare, finance, transportation, and energy, where time-series data are fundamental components. The current works are still limited in their ability to perform reasoning that involves both time-series and the corresponding textual content. We address this gap by introducing \textit{Chat-TS}, a large language model (LLM) based framework designed to support reasoning over time series and textual data. Unlike traditional models, Chat-TS integrates time-series tokens into LLMs' vocabulary, enhancing its reasoning ability over both modalities without compromising core natural language capabilities. To support learning and evaluation, we contribute new datasets: the \textit{TS Instruct Training Dataset} (pairing diverse time-series data with relevant text instructions and responses for instruction tuning), the \textit{TS Instruct Question and Answer (QA) Gold Dataset} (multiple-choice questions to evaluate multimodal reasoning), and a \textit{TS Instruct Quantitative Probing Set} (a small subset of TS Instruct QA reasoning tasks alongside math and decision-making questions for LLM evaluation). We design a training strategy to preserve the inherent reasoning capabilities of LLMs while augmenting them for time-series reasoning. Experiments show that Chat-TS achieves state-of-the-art performance in multimodal reasoning tasks by maintaining strong natural language proficiency while improving time-series reasoning.\footnote{To ensure replicability and facilitate future research, all models, datasets, and code will be available at \scriptsize\texttt{https://github.com/quinlanp/Chat-TS-Multi-Modal-Reasoning}.}
\end{abstract}

\section{Introduction}
Large language models (LLMs) are rapidly being applied in many fields which necessitate the analysis of time-series such as healthcare~\citep{10.1145/3531326}, business~\citep{https://doi.org/10.1002/widm.1519}, transportation~\citep{zhang2024largelanguagemodelsmobility}, and many other industries~\citep{app132212374}. Traditional time-series analysis tasks, such as forecasting, imputation, anomaly detection, and classification, have been extensively studied in pure time-series setups~\citep{spyros_makridakis__1977, Dempster_2020,jiang2024empoweringtimeseriesanalysis}. The most recent advancements in large language models(LLMs)~\citep{openai2024gpt4technicalreport,geminiteam2024geminifamilyhighlycapable,dubey2024llama3herdmodels,abdin2024phi3technicalreporthighly} have opened significant opportunities for performing time-series analysis in the form of textual reasoning in a multimodal setup. The practical application of time-series data analysis often necessitates drawing conclusions and making decisions based on real-world knowledge, which is conveyed and stored in human language. For example, an investment banker may consider both stock prices and relevant company reports, while a doctor may analyze a patient's ECG readings in combination with their medical history and clinical notes.

Recent research has started to apply large language models to time-series analysis, e.g., as financial trading agents~\citep{wang2023alphagpthumanaiinteractivealpha,wang2024llmfactorextractingprofitablefactors,li2023tradinggptmultiagentlayeredmemory} and medical analysts~\citep{informatics11030057,Thirunavukarasu2023}. The models are applied within their specific domains and time-series are first converted from numerical representations to a textual format in order to leverage LLMs. This approach has also been used in pure time-series analysis (e.g., forecasting) by~\cite{gruver2024largelanguagemodelszeroshot}. These models serve as a state-of-the-art baseline for comparing language model reasoning over time-series. A comprehensive discussion on previous work will be detailed in Section \ref{sec:related}. Despite the promising applications of LLMs to handle real-world multi-domain challenges, the training and evaluation of LLMs' reasoning abilities for time-series data remains significantly under-explored.

Developing LLMs with time-series (TS) reasoning capabilities presents several key challenges: (1) Training data scarcity: There is a significant lack of training data for constructing models that combine time-series data with reasoning. Multimodal datasets that integrate both text and time-series data are currently limited. (2) Benchmarking: Similarly, there are no large-scale benchmarks available to evaluate multimodal reasoning and decision-making for time-series data. (3) Preserving LLM Capabilities: Current time-series approaches that rely on LLMs for time-series reasoning either do not preserve natural language knowledge and reasoning capabilities inherent in LLMs or do not evaluate it. We advocate that this should be standard in future multi-modal reasoning works.
To address these challenges, our paper makes the following contributions:

\begin{itemize}[leftmargin=*]
    \item\textbf{We contribute new datasets} for training and evaluating LLMs for time-series reasoning tasks. The \textbf{TS Instruct Training Dataset}, a diverse multimodal dataset that integrates time-series data with relevant textual information, fills a critical gap in the current literature by addressing the scarcity of training datasets. The \textbf{TS Instruct QA Gold Benchmark}, a dataset of 1,056 multiple-choice questions with ground truth answers, is designed to evaluate multimodal reasoning capabilities in time-series analysis. We also provide a small dataset for quantitative evaluation, composed of QA, mathematical reasoning, and decision-making samples.
    \item\textbf{We propose an effective approach} to leveraging multimodal LLMs for time-series analysis with text, emphasizing the maintenance of LLMs’ reasoning capabilities and using their knowledge reservoir to enhance time-series data reasoning. We expand the embedding space while preserving the LLM's capabilities, enabling strong reasoning after fine-tuning with multimodal data, outperforming existing methods that compromise natural language and reasoning ability.
    \item\textbf{Experiments demonstrate} that the proposed Chat-TS Models improve time-series reasoning capabilities by an average of $\sim$13\% on the TS Instruct QA Gold Benchmark, compared to the existing state-of-the-art baseline.
\end{itemize}


\section{Related Work}
\label{sec:related}

\textbf{LLMs for Time-Series Analysis.} Many recent works treat large language models as a backbone for forecasting or classification through repurposing their pretrained parameters, but at the cost of disabling their full text‐generation and reasoning abilities. For instance, GPT4TS~\citep{zhou2023fitsallpowergeneraltime} and LLM4TS~\citep{chang2024llm4tsaligningpretrainedllms} embed series in model weights; TEMPO~\citep{cao2024tempopromptbasedgenerativepretrained}, and TEST~\citep{sun2024testtextprototypealigned} insert series into prompts; and InstructTime~\citep{cheng2024advancingtimeseriesclassification} fine-tunes for classification, without preserving general multimodal reasoning.
LLMTime~\citep{gruver2024largelanguagemodelszeroshot} instead encodes time-series as text, enabling zero-shot forecasting without task-specific fine-tuning and retaining the model’s text-based structure. 

\paragraph{Time-Series Reasoning with LLMs.} A few recent efforts have begun to explore time-series reasoning as a multi-modal natural language task as opposed to the traditional time-series analysis tasks mentioned above. ChatTime~\citep{chattime} and TimeMQA~\citep{timemqa} build generated datasets which test time-series alignment (eg. identifying trends) with LLMs but train and test on subsets from the same global datasets which limit comparability. Chow et al.~\citeyear{chow2024timeseriesreasoningllms} augment an LLM with a learned encoder for raw series inputs. However, neither their code nor the assessment of the model’s natural-language capabilities has been released. In contrast, Chat-TS not only preserves end-to-end text input and generation, but shows strong generalization on our TS-Instruct QA Gold dataset. This dataset focuses on real-world reasoning scenarios, despite only training on simple conversational samples. We publish both model and dataset benchmarks. A concurrent work \citep{xie2025chattsaligningtimeseries} trains LLMs for time-series reasoning. This work uses a trained encoder and embedding concatenation to combine time-series representations. We believe our methods are orthogonal to~\citep{xie2025chattsaligningtimeseries}. For details, please refer to Appendix~\ref{APPENDIX:vision_models}.

\paragraph{Time-Series Training and Evaluation.} Standard repositories (Monash~\citep{godahewa2021monash}, UCR~\citep{Middlehurst_2024}, LOTSA~\citep{woo2024unifiedtraininguniversaltime}) offer forecasting and classification benchmarks but lack textual context for multimodal reasoning. Merrill et al.~\citeyear{merrill2024languagemodelsstrugglezeroshot} evaluate LLM reasoning on synthetic series and LLM‐generated prompts, highlighting performance gaps but without providing training improvement strategies. We address this gap by releasing (1) a large-scale conversational training corpus of human-LLM exchanges grounded in real-world series metadata, and (2) TS-Instruct QA Gold, a human-validated question–answer set over real series with visual context. Merrill also contributed a single-time-series question and answer (QA) based dataset using synthetic dataset. Unfortunately they deemed it not appropriate for LLM evaluation as existing text models had near perfect performance on these benchmarks. In contrast we contribute a challenging QA benchmark which consists of real-world time-series and utilize human evaluators to verify the accuracy of each question.

\paragraph{Discrete Tokenization.} Existing multimodal tokenizers (e.g., VQ-VAE~\citep{oord2018neuraldiscreterepresentationlearning} or channel-wise patching) optimize for compression on very long sequences, often degrading reconstruction fidelity. We introduce a lightweight discrete tokenizer aimed at maximizing accuracy on sub-1,000-point series. While Chronos~\citep{ansari2024chronoslearninglanguagetime} uses similar discrete tokenization in LLM-adjacent models, it does not maintain full LLM text-generation or reasoning capabilities and focuses on time-series forecasting.

\section{A Framework for Reasoning Over Time-Series and Language Data}

\begingroup
\setlength{\dbltextfloatsep}{4pt}
\setlength{\abovecaptionskip}{1pt}
\setlength{\belowcaptionskip}{0pt}

\begin{figure*}[t]
\centering

\begin{minipage}[t]{.48\textwidth}
\centering\vspace{0pt}%
\resizebox{\linewidth}{!}{%
\begin{tikzpicture}[
  baseline=(current bounding box.north), 
  font=\footnotesize, >=Latex, every node/.style={transform shape},
  arrow/.style={-Latex, thick, shorten <=1.2pt}, 
  box/.style={draw, rounded corners, align=center, inner sep=2.1pt, minimum height=5.0mm},
  token/.style={draw, rounded corners=1pt, inner sep=1.0pt, font=\scriptsize, fill=white},
  data/.style={box, fill=blue!7, minimum width=27mm},
  tbox/.style={box, fill=orange!10, minimum width=28mm},
  vbox/.style={box, fill=green!10, minimum width=32mm}
]
\newcommand{\icon}[1]{\raisebox{-0.12em}{\faIcon{#1}}\,}
\path[use as bounding box] (0,0) rectangle (10,4.85); 

\node[data] (textin) at (3.2,0.45) {\icon{align-left}\ \textbf{Text input}};
\node[data] (tsin)   at (6.8,0.45) {\icon{chart-line}\ \textbf{Time-series input}};

\node[tbox] (txttok) at (3.2,1.40) {\icon{font}\ Text tokenizer\\ \scriptsize outputs $\mathcal{V}_L$};
\node[tbox] (tstok)  at (6.8,1.40) {\icon{cogs}\ TS tokenizer\\ \scriptsize bins+$T_c$ $\rightarrow \mathcal{V}_T$};

\node[token] (ltoks) at (3.2,2.30) {\scriptsize text tokens $\mathcal{V}_L$};
\node[token] (ttoks) at (6.8,2.30) {\scriptsize TS tokens $\mathcal{V}_T$};

\node[vbox] (joint)  at (5.0,3.55) {\icon{boxes}\ Joint token stream\\ $\mathcal{V}=\mathcal{V}_L\cup\mathcal{V}_T$};

\draw[arrow] (textin.north) -- (txttok.south);
\draw[arrow] (tsin.north)   -- (tstok.south);
\draw[arrow] (txttok.north) -- (ltoks.south);
\draw[arrow] (tstok.north)  -- (ttoks.south);
\draw[arrow,shorten >=1.6pt] (ltoks.north) -- ++(0,0.40) -| (joint.south);
\draw[arrow,shorten >=1.6pt] (ttoks.north) -- ++(0,0.40) -| (joint.south);

\end{tikzpicture}%
}
\end{minipage}
\hfill
\begin{minipage}[t]{.48\textwidth}
\centering\vspace{0pt}%
\resizebox{\linewidth}{!}{%
\begin{tikzpicture}[
  font=\footnotesize, >=Latex, every node/.style={transform shape},
  arrow/.style={-Latex, thick},
  layer/.style={draw, rounded corners, align=center, inner sep=2pt, minimum height=5.0mm, minimum width=30mm},
  trainable/.style={layer, fill=green!12, draw=green!50!black},
  frozen/.style={layer, fill=gray!18, draw=gray!60},
  token/.style={draw, rounded corners=1pt, inner sep=1.0pt, font=\scriptsize, fill=white},
  phase/.style={draw, rounded corners, inner sep=2.5pt}
]
\newcommand{\icon}[1]{\raisebox{-0.12em}{\faIcon{#1}}\,}
\path[use as bounding box] (0,0) rectangle (10,4.85); 

\node[font=\scriptsize, align=center, inner sep=0.8pt] (leg) at (5,4.60)
{{\color{green!50!black}\rule{7pt}{6pt}} trainable \quad {\color{gray!60}\rule{7pt}{6pt}} frozen};

\node[font=\scriptsize\bfseries] (p1title) at (2.7,4.00) {Phase 1: TS-only pretrain};
\node[token]     (p1_tok)  at (2.7,0.60)  {TS $\mathcal{V}_T$};
\node[trainable] (p1_emb)  at (2.7,1.40)  {\icon{font}\ Embeddings ($\mathcal{V}$)};
\node[frozen]    (p1_blk)  at (2.7,2.20)  {\icon{robot}\ Transformer blocks};
\node[trainable] (p1_head) at (2.7,3.00)  {\icon{dice-d6}\ LM head};
\draw[arrow] (p1_tok.north) -- (p1_emb.south);
\draw[arrow] (p1_emb.north) -- (p1_blk.south);
\draw[arrow] (p1_blk.north) -- (p1_head.south);
\node[phase, fit=(p1title)(p1_tok)(p1_head)] {};

\node[font=\scriptsize\bfseries] (p2title) at (7.3,4.00) {Phase 2: Instruction tuning (text + TS)};
\node[token]     (p2_ltok)  at (6.5,0.60)  {text $\mathcal{V}_L$};
\node[token]     (p2_ttok)  at (8.1,0.60)  {TS $\mathcal{V}_T$};
\node[trainable] (p2_emb)   at (7.3,1.40)  {\icon{font}\ Embeddings ($\mathcal{V}$)};

\coordinate (p2_merge) at (7.3,0.98);          
\draw[thick] (p2_ltok.north) -- ++(0,0.18) -| (p2_merge);
\draw[thick] (p2_ttok.north) -- ++(0,0.18) -| (p2_merge);
\draw[arrow] (p2_merge) -- (p2_emb.south);     

\node[trainable] (p2_blk)   at (7.3,2.20)  {\icon{robot}\ Transformer blocks};
\node[trainable] (p2_head)  at (7.3,3.00)  {\icon{dice-d6}\ LM head};
\draw[arrow] (p2_emb.north) -- (p2_blk.south);
\draw[arrow] (p2_blk.north) -- (p2_head.south);
\node[phase, fit=(p2title)(p2_ltok)(p2_head)] {};

\end{tikzpicture}%
}
\end{minipage}

\caption{\textbf{Left:} Text and time-series inputs are tokenized by their respective tokenizers and merged into a joint token stream over the extended vocabulary $\mathcal{V}$. \textbf{Right:} Chat-TS training. \emph{Phase 1} pretrains on TS tokens with frozen transformer blocks; \emph{Phase 2} instruction-tunes on both text and TS tokens.}
\label{fig:overview_fig}
\end{figure*}
\endgroup


\subsection{Problem Definition}
This paper is concerned with multimodal time-series reasoning involving a model's ability to make decisions and answer relevant questions based on time-series, alongside the corresponding text data.
An key objective of the design is to expand the vocabulary of any pre-trained LLM to include time-series tokens, training autoregressively along with text. This strategy aims to augment the quality of time-series understanding and reasoning while maintaining the original natural language understanding (NLU) capabilities of a large language model. 

Formally, let $\mathcal{V}_L$ denote the original textual vocabulary and $\mathcal{V}_T$ the set of time-series tokens, forming an extended vocabulary $\mathcal{V} = \mathcal{V}_L \cup \mathcal{V}_T$. The LLM, parameterized by $\theta$, is trained to predict the next token in a sequence $\mathbf{y} = \{y_1, y_2, \dots, y_{T}\}$, where each $y_t \in \mathcal{V}$, using the autoregressive model $f_\theta: \mathcal{V} \to \mathcal{V}$. Following the convention, the training process maximizes the likelihood of the next token via cross-entropy (CE) loss: $\mathcal{L}(\theta) = -\sum_{t=1}^{T-1} \log p_\theta(y_{t+1} | y_1, y_2, \dots, y_t)$, where $p_\theta$ is the predicted probability of the next token. Since $\mathcal{V}$ contains joint representations for text and time-series, the training process is reduced to the standard auto-regressive LLM training.


\subsection{Main Components}
The key focus of developing our joint text and time-series models lies in the creation of training data, the development of tokenization methods for time series, and their integration into LLMs for multimodal time-series analysis. A high-level overview is shown in Figure~\ref{fig:overview_fig}.

Data is the cornerstone of any multimodal learning approach, and high-quality text and time-series data are critical for training robust models. However, while there are many high-quality datasets available for either text or time-series, there are no existing datasets that integrate both modalities---a significant barrier to instruction-tuning LLMs for time-series tasks. To address this, we contribute novel datasets designed and curated specifically for this purpose, using a combination of time-series properties, task descriptors, and synthetic conversations, as will be detailed later in this section.

Our framework enables learning over both text and time-series data, in order to allow the model to reason over time-series data while preserving its original language modeling capabilities. Unlike other multimodal models that require connectors to modify representations for LLM integration~\citep{mckinzie2024mm1methodsanalysis}, we advocate adding time-series tokens to the LLM’s vocabulary, eliminating the need for intermediary connecting components between the time-series and text representation spaces and therefore ensuring a more seamless integration. Concretely, we explore two initialization methods: (i) mean initialization, where embeddings are initialized with the average of existing text tokens, and (ii) time-series pre-training, where we unfreeze and tune only the embedding and final linear layer.

We then fine-tuned the models using a dual-dataset strategy that combines instruction-answer pairs from our multimodal TS-Instruct dataset together with a curated subset of a text-based instruction tuning dataset. This approach preserves the LLMs' strong language modeling capabilities while enhancing their power to follow instructions and reason effectively over time-series data. Below, we discuss the details of the components.

\subsection{Time-Series-with-Text Multimodal Dataset Curation}
\label{dataset_synth}

Within this work we generate datasets for training and testing using real non-overlapping time-series. To prevent contamination between datasets we take several important measures. During training we only generate simple single-turn casual conversations between a user and assistant. We provide some examples in Appendix \ref{APPENDIX: Dataset Generation}. Conversely, for the evaluation datasets, each sample is generated in the multi-choice question-answer format with much stricter conditions.

Importantly, there are a few recent works which build time-series reasoning datasets in some capacity. For example ChatTime\citep{chattime} and Time-MQA\citep{timemqa}. These datasets in terms of reasoning primarily focus on time-series alignment (eg. Identify the trend), where as ours is concerned with real-world reasoning questions, but they first generate their dataset and then split it into testing and training. In contrast, both our training and evaluation datasets are generated separately in different formats to more broadly evaluate the models generalization capabilities through this controlled separation.  

To highlight that our model performance not only improves on our benchmark we also compare to the two time-series question answering dataset (MCQ2TS) as presented \cite{merrill2024languagemodelsstrugglezeroshot}. This is a time-series reasoning dataset that is based off of synthetically generated time-series and requires counter-factual reasoning over two time-series and is remarkably distinct from all datasets generation within this work.

We curated new datasets to help address the critical scarcity of multimodal training and benchmarking test data for time-series reasoning. As detailed in Appendix~\ref{appendix:assets}, all datasets will be made publicly available. 




\paragraph{The TS-Instruct Dataset.} As there is limited multi-modality time-series data available, we synthesize conversations between a user and an assistant using GPT4o-mini~\citep{openai2024gpt4technicalreport}. These dialogues incorporate time-series data represented as images alongside metadata (length, channels) to guide the model and generate accurate conversations with real-world time-series for model training. Conversations are designed to focus on general reasoning, classification, decision-making, and mathematical analysis. Time-series data for reasoning tasks are sourced from LOTSA~\citep{woo2024unifiedtraininguniversaltime}, while classification labels come from the Time-Series Classification Archive~\citep{Middlehurst_2024}. The details of the dataset can be found in Appendix~\ref{APPENDIX: Dataset Generation}.
\paragraph{The TS-Instruct QA Gold Benchmark.} We designed a multimodal evaluation dataset to address the lack of benchmarks for multimodal models capable of time-series reasoning. The dataset consists of multiple-choice question-and-answer (QA) tasks, where each question includes a description of the time-series data, four answer options, and a detailed explanation of the correct answer. This allows us to assess both answer accuracy and reasoning ability. The dataset was generated using GPT-4o-mini, which, combined with visual representations and time-series meta-data is used to generate the text components of the multiple-choice questions. The full system prompt is detailed in Appendix~\ref{APPENDIX: System_Prompts}.

To ensure the quality of the evaluation data, each sample was manually evaluated by hand for quality and to ensure each samples had a single correct answer. For more details see Appendix \ref{App:mturk}.
The Gold dataset serves as a reliable benchmark for time-series reasoning under optimal conditions, providing a fair basis for comparing models. After pruning, the TS Instruct QA Gold benchmark consists of 1,056 samples. The correct answers are balanced so that each option appears 264 times within the dataset.

\subsection{The Tokenization Model} 
\label{sec:discrete_tok}
In order to preserve natural language reasoning we aim to expand the embedding vocabulary space in combination with a time-series tokenizer as a minimally invasive method for incorporating time-series reasoning.
Tokenizing time-series data presents unique challenges, as it involves converting continuous numerical values into discrete tokens that can be processed by an LLM. Our approach balances reconstruction accuracy with token scalability, ensuring that the tokenized time-series can be efficiently represented without compromising data integrity.

There are several key considerations when designing our time-series tokenizer:
\begin{enumerate}\setlength\itemsep{0pt}\setlength\parskip{0pt}\setlength\parsep{0pt}
    \item \textbf{Reconstruction accuracy}: The tokenizer should encode and decode time-series data with minimal error.
    \item \textbf{Compression}: It should allow for efficient compression to process larger time-series sequences.
    \item \textbf{Scalability}: The tokenizer needs to support varying vocabulary sizes, improving quality as the number of tokens increases.
\end{enumerate}

Our discrete tokenizer functions as follows: Let \( \bm{X} \in \mathbb{R}^{L \times M} \) represent a multivariate time-series, where \( L \) is the sequence length and \( M \) is the number of channels. We quantize the time-series by dividing the range \( [-s, s] \) into \( K-1 \) bins, mapping each value \( x \) to its corresponding bin index \( i \), which forms a sequence of discrete tokens \( \bm{Z} \). A special token \( T_c = K\) is added to mark the end of each channel, and the token sequence \( \bm{T} \) is then flattened and offset to align with the vocabulary size of the text tokenizer. Decoding reverses this process, reconstructing the time-series by mapping each index back to the bin centers. The special token is used as a channel separator to reconstruct the original 2D time-series from the 1D sequence of tokens. We use \(s=\pm3\sigma\) of the normalized time-series and a codebook size \(K\) to be 8192 tokens.

We implemented and evaluated four tokenizers, including models using simple linear layers for encoding/decoding with vector quantization (VQ)~\citep{oord2018neuraldiscreterepresentationlearning}, lookup-free quantization (LFQ)~\citep{yu2024language}, and a TCN-based~\citep{oord2016wavenetgenerativemodelraw} tokenizer using residual vector quantization (RVQ)~\citep{Chen2010} as implemented in InstructTime~\citep{cheng2024advancingtimeseriesclassification}. We compare these approaches to our  discrete tokenizer that requires no training and provides near-perfect reconstruction of time-series data.

\begin{figure}[h]
  \centering
  \includegraphics[width=\linewidth]{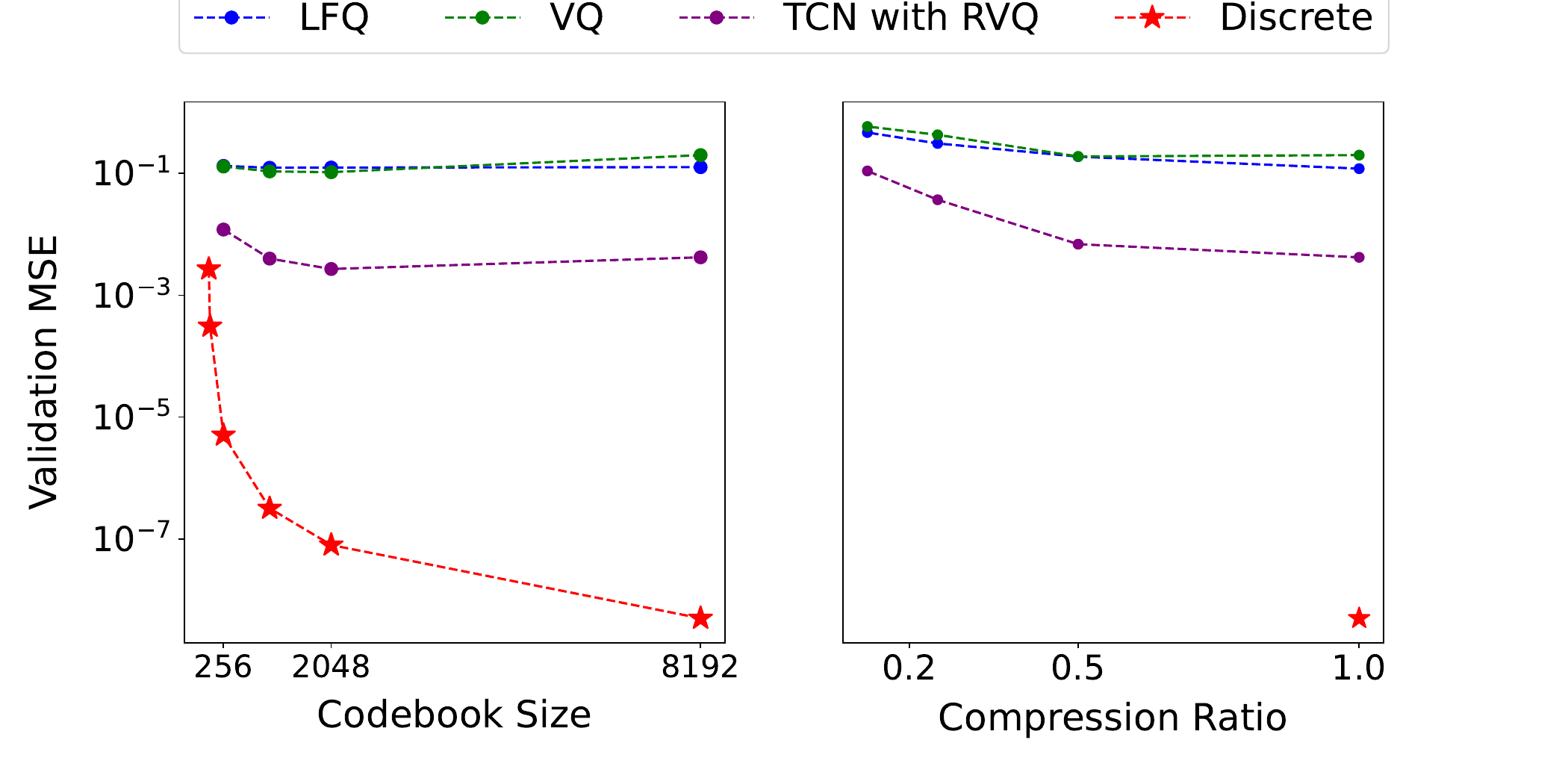}
  \caption{Tokenizer validation: reconstruction error under varying compression ratio and codebook size .}
  \label{fig:tok_comparison}
\end{figure}

\paragraph{Tokenizer Selection.}
There are many options in terms of time-series tokenizer architecture (simple discrete, vector quantization based etc.) which would work within our framework. To select the best tokenizer, 
we measured reconstruction quality using varying compression ratios and codebook sizes on the Azure VM Traces 2017 dataset~\citep{hadary2020protean}. Figure \ref{fig:tok_comparison} presents the comparison, showing that our discrete tokenizer outperforms the others in terms of validation error and improves predictably as the codebook size increases. 

The results demonstrate that the discrete tokenizer significantly outperforms the other tokenizers in terms of validation error and predictably improves when scaling the number of codebook tokens. Importantly, our tokenizer does not require any training and therefore does not suffer from out of distribution problems that may arise otherwise. It is important to note that regardless of choice in tokenizer setup (simple discrete, vq-based etc.) is considerably \textbf{compressed} compared to the baseline text encoding strategies used in modern LLMs.

\section{Experimental Setup}
We use different types of datasets for pretraining, instruction tuning, and model evaluation. Full details on each data type can be found in Appendix~\ref{APPENDIX: Dataset Generation}.

\paragraph{Embedding Initialization and Pretraining.}  
 We test two methods for integrating time-series tokens: mean initialization, and pre-training with time-series data. Mean initialization sets new time-series embeddings based on the average of text tokens in the model vocabulary, while pre-training unfreezes a small portion of the model (embedding layer and final linear layer) to allow these new embeddings to specialize. Freezing most of the model preserves the LLM's existing capabilities while allowing the new embeddings to learn from time-series data. Next, each model is trained on our TS Instruct dataset and the Open Orca text dataset based on the specific setups.
 
We utilize the LOTSA~\citep{woo2024unifiedtraininguniversaltime} dataset for model pretraining, which encompasses a wide variety of time-series domains, providing diverse sequence lengths and channel dimensions. Time-series samples are generated through a sliding window approach with configurable parameters for the window size, stride, and channel selection. This method maximizes sample diversity and sizes, ensuring broad coverage of temporal patterns.

\paragraph{Instruction Tuning.}  
To enhance and sustain performance on natural language tasks, we incorporate 100,000 samples from Open-Orca~\citep{OpenOrca}, a high-quality source of diverse natural language instruction data. During instruction tuning, we blend this natural language data with the TS-Instruct multimodal dataset. This dual-dataset integration increases multimodal reasoning capabilities of the models while preserving their core language reasoning strengths.

\paragraph{Multimodal Reasoning Evaluation.}  
We conduct a comprehensive evaluation and internal ablation studies to assess the impact of training data on performance. We evaluate the models using the TS-Instruct QA Gold Dataset and its subsets for quantitative analysis. Following \citep{merrill2024languagemodelsstrugglezeroshot,xie2025chattsaligningtimeseries,chow2024timeseriesreasoningllms}, and to maintain comparability with existing literature, we use the LLMTime~\citep{gruver2024largelanguagemodelszeroshot} text encoding method for all multi-modal reasoning baselines and compare directly to the baseline. This method allows us to make direct comparisons in reasoning performance when comparing the trained Chat-TS model to its original core architecture and test across many different baseline models. 


\paragraph{Training Details}
In our experiments we use the 8 billion parameter LLama 3.1 model~\citep{dubey2024llama3herdmodels} which provides a reasonable balance between performance and compute. Our pretraining stage trains on 115,000 time-series samples for 1 epoch using an initial learning rate of $2e-3$. We then instruction tune the model on our dataset comprising roughly 18,000 multimodal samples and 100,000 text samples with a learning rate of $2e-5$ for 1 epoch. We use cosine-scheduling for the learning rates. For more information refer to Appendix~\ref{APPENDIX:Training_Details}.

\section{Experiment Results and Analysis}
\label{sec:results}
Our experiment results and analysis are organized from four perspectives: (1) \textbf{Baseline Comparison} on the Gold dataset to benchmark reasoning capabilities under controlled conditions; (2) \textbf{Natural Language Performance} that evaluates the impact of time-series instruction tuning on the model's general language understanding and reasoning abilities; (3) \textbf{Training Data Composition} evaluates several variants of the model using different combinations of training data; and (4) \textbf{Quantitative Analysis of Generated Explanations} assess the quality of responses using metrics such as helpfulness, relevance, accuracy, and level of detail.






\subsection{Baseline Comparison on TS-Instruct QA Gold}
\label{sec:baseline_comparison}

We compared our model against baselines using the curated \textbf{TS-Instruct QA Gold} dataset. Table~\ref{tab:ts_vs_mmlu} presents the detailed results, ranking models by TS reasoning performance and using MMLU-Pro scores~\citep{wang2024mmluprorobustchallengingmultitask} to assess general NLP capabilities.

\begin{table}[h]
\centering
\scriptsize
\begin{tabular}{lccc}
\toprule
\textbf{Model} & \textbf{TS Perf.} & \textbf{MMLU-Pro} & \textbf{Params (B)} \\
\midrule
Gemma-2 2B & 42.18 & 0.2719 & 2 \\
Gemma-2 9B & 49.64 & 0.4125 & 9 \\
LLama 3.1-8B(base) & 54.22 & 0.3772 & 8 \\
Ministral-8B & 58.72 & 0.3483 & 8 \\
Phi-3-medium-4k & 64.64 & 0.474 & 14 \\
\textbf{Chat-TS (ours)} & 67.22 & 0.356 & 8 \\
GPT-4o-mini$^*$  & 74.55 & 0.631 & unknown \\
GPT-4o$^*$  & 78.50 & 0.766 & unknown \\
\bottomrule
\end{tabular}
\caption{Comparison of TS Reasoning and NLP (MMLU-Pro) performance. Models are ranked by TS Reasoning performance. Baselines include Gemma-2 2B~\citep{gemmateam2024gemmaopenmodelsbased}, Gemma-2 9B~\citep{gemmateam2024gemmaopenmodelsbased}, Llama 3.1-8B~\citep{dubey2024llama3herdmodels} the base model used for Chat-TS, Ministral-8B~\citep{jiang2023mistral7b}, Phi-3-medium-4k~\citep{abdin2024phi3technicalreporthighly}, GPT-4o-mini and GPT-4o~\citep{openai2024gpt4technicalreport}. $^*$ Indicates that this family of models was used in dataset generation.}
\label{tab:ts_vs_mmlu}
\end{table} 

 The results show the importance of both model sizes and time-series specific training. Despite the  size of our model, it outperforms models such as Phi-3-medium-4k-instruct which are both larger and has strong reasoning capabilities on NLP benchmarks. These results also show the importance of scale, since scaling up model size/performance (for example gemma-2b to gemma-9b) generally increases performance on both the MMLU-Pro and time-series benchmarks. For reference, we included into our analysis the state-of-the-art closed-source models such as GPT-4o and GPT-4o-mini, although these results should be interpreted with caution as the GPT-4o-mini (which likely shares training data with GPT-4o) was used in the original dataset generation process.


To show that the increased reasoning performance of Chat-TS compared to its base LLM, we also compare with 10,000 samples from the MCQ-2TS dataset from \cite{merrill2024languagemodelsstrugglezeroshot}. This dataset is dramatically different in design compared to anything our model has seen during training. This dataset consists of questions which test counterfactual reasoning across two time-series. Comparatively, our model is only trained on casual single-turn conversations with singular time-series. Chat-TS shows a substantial increase in performance on this reasoning task as shown in Table \ref{mcq2ts} indicating performance gains are not due to dataset contamination of the generated datasets.

\begin{table}[h]
\centering
\scriptsize
\begin{tabular}{l c}
\toprule
\textbf{Model} & \textbf{MCQ2TS Score (\%)} \\
\midrule
Llama-3 8B & 36.5 \\
Chat-TS (PreOrcaTS) & 47.6 \\
\bottomrule
\end{tabular}
\caption{MCQ2TS benchmark results comparing Llama-3 8B and Chat-TS (PreOrcaTS).}
\label{mcq2ts}
\vspace{-2mm}
\end{table}

\subsection{Model Ablations and Analysis}
\label{Model_Val}
We study the contributions of each training component. This includes examining (i) the impact of using only text data from the Open Orca dataset, (ii) the effect of pretraining the added time-series tokens using the time-series dataset described in Appendix \ref{APPENDIX: Dataset Generation}, and (iii) the role of instruction-tuning with the TS Instruct dataset. The abbreviations in Table \ref{model_abrv} correspond to each model variant and its training data.

\begin{table}[h]
\small
\centering
\resizebox{.99\columnwidth}{!}{%
\begin{tabular}{ll}
\toprule
\textbf{Chat-TS Variant} & \textbf{Training Components} \\
\midrule
{Orca} & LLama 3.1-8B + Open Orca (Text only) \\
{OrcaTS} & LLama 3.1-8B + Open Orca + TS Instruct \\
{TS} & LLama 3.1-8B + TS Instruct \\
{PreTS} & LLama 3.1-8B + TS Pre-Training + TS Instruct \\
\textbf{PreOrcaTS(Chat-TS)} & LLama 3.1-8B + Open Orca + TS Pre-Training + TS Instruct \\
\bottomrule
\end{tabular}
}
\caption{Model Variants and Training Components. We use PreOrcaTS as the complete Chat-TS model.}
\label{model_abrv}
\end{table}

\subsubsection{Preserving NLP Capabilities in Multimodal Models}
\label{sec:preserving_nlp_capabilities}

\begin{table}[tb]
\centering
\resizebox{.99\columnwidth}{!}{%
\scriptsize
\begin{tabular}{lccc}
\toprule
\textbf{Training Data} & \textbf{MMLU-Pro} & \textbf{Big-Bench-Hard} & \textbf{GPQA}\\
\midrule
Llama-3.1-8B  & $0.376 \pm 0.004$ & $0.511 \pm 0.154$ & $0.324 \pm 0.029$\\
Orca          & $0.367 \pm 0.004$ & $0.529 \pm 0.156$ & $0.328 \pm 0.035$\\
OrcaTS        & $0.356 \pm 0.004$ & $0.522 \pm 0.158$ & $0.341 \pm 0.042$\\
TS            & $0.360 \pm 0.004$ & $0.530 \pm 0.161$ & $0.322 \pm 0.032$\\
PreTS         & $0.363 \pm 0.004$ & $0.526 \pm 0.158$ & $0.318 \pm 0.014$\\
PreOrcaTS     & $0.356 \pm 0.004$ & $0.525 \pm 0.155$ & $0.340 \pm 0.042$\\
\bottomrule
\end{tabular}
}
\caption{Accuracy of Llama-3.1-8B variants on MMLU-Pro, Big-Bench-Hard, and GPQA.
All scores are within $2\%$ of one another.}
\label{tab:mmlu_comparison}
\vspace{-2mm}
\end{table}
In this section we show that the Chat-TS architecture preserves the natural language capabilities of the core large language model.
Previous research \citep{zhou2023fitsallpowergeneraltime,chang2024llm4tsaligningpretrainedllms} has integrated LLMs as components of time-series models. However, a common issue with these models is that they often fail to retain the general natural language understanding and reasoning capabilities of the underlying LLMs. 

Table~\ref{tab:mmlu_comparison} presents the performance of different model variants across three natural language benchmarks: MMLU-Pro \citep{wang2024mmluprorobustchallengingmultitask}, Big Bench Hard \citep{suzgun2022challengingbigbenchtaskschainofthought}, and GPQA \citep{rein2023gpqagraduatelevelgoogleproofqa}. The results demonstrate that our multimodal LLMs maintain comparable performance to the base model, indicating that integrating time-series reasoning does not degrade the model's natural language understanding capabilities.
\begin{figure*}[h]
    \centering
    \includegraphics[width=0.8 \textwidth]{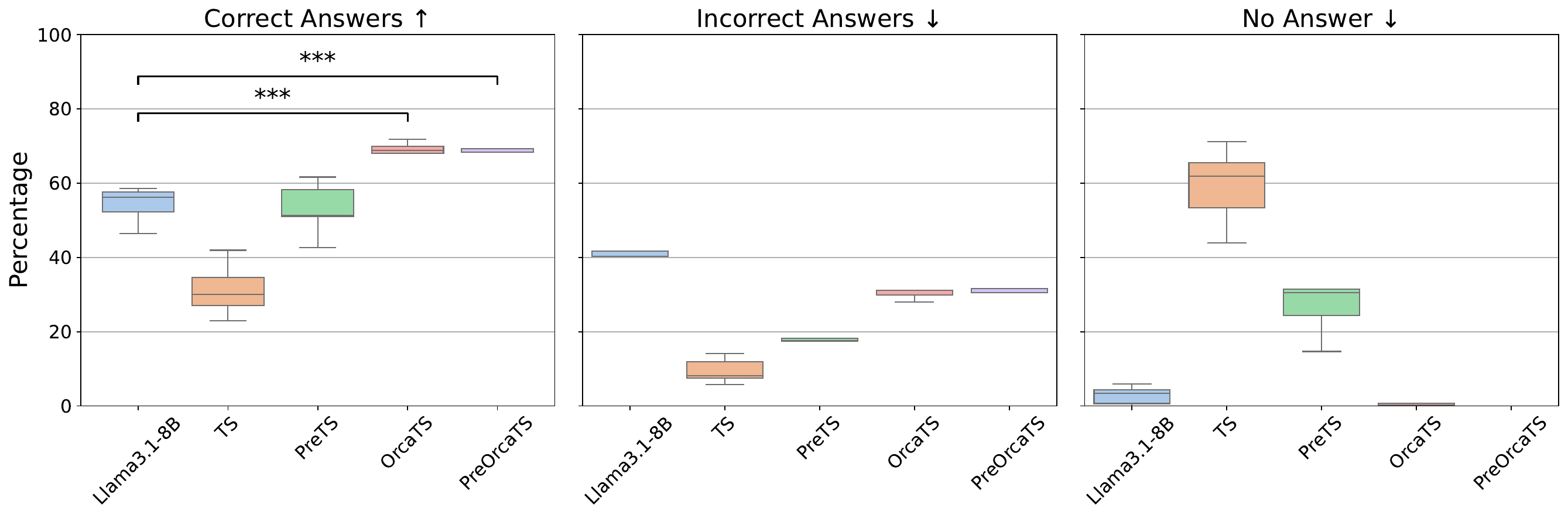}
    \caption{Model ablation on the TS-Instruct QA Gold  dataset with Chat-TS trained on different data combinations~(Table \ref{model_abrv}). Statistical significance is indicated by *** (p-value $<$ 0.05).}
    \label{QA_results}
\end{figure*}


We can see that even the models trained \textbf{solely on the TS Instruct Dataset} (i.e., without Open Orca data) achieve performance nearly identical to the models trained with interleaved text and time-series data. This stands in contrast to observations in other multimodal domains, such as computer vision, where models often rely on trained connectors to link visual and textual representations \citep{mckinzie2024mm1methodsanalysis}. In such domains, text data remains imperative for preserving the underlying LLM capabilities. Each model is within $\pm2\%$ of the original base LLama 3.1-8B model. These differences can be attributed to the differences in alignment from the full parameter fine-tuning and does not necessarily indicate a increase or decrease in overall model quality.

The results demonstrate that expanding the model’s vocabulary with discrete tokenizers is an effective way to extend LLM capabilities without compromising natural language reasoning performance. This ensures that the model retains its foundational NLP strengths while acquiring the ability to reason effectively with time-series data.


\subsubsection{TS-Instruct QA Gold Ablation}
\label{sec:ts_instruct_qa}

To evaluate performance at scale, we use the TS-Instruct QA Gold dataset. This larger evaluation allows us to assess robustness and instruction-following capabilities across varied prompt configurations. The evaluation was repeated 5 times with different system prompts which can be found in Appendix \ref{APPENDIX: System_Prompts}. Figure~\ref{QA_results} demonstrates that PreOrcaTS and OrcaTS (using mean initialization) consistently outperforms other model variants, including the current state-of-the-art time-series encoding strategy, in both accuracy and robustness to prompt variability. Models trained solely on time-series data (TS, PreTS) struggle to produce correct answers, primarily due to their difficulty in adhering to instruction formats. This highlights the importance of interleaving text and time-series data during training particularly for multimodal analysis.

\subsubsection{Quantitative Analysis of Text Quality}
\label{sec:quantitative_analysis}

In this section, we present a quantitative analysis of each model variant using GPT-4o as the evaluator. While the full TS-Instruct QA Gold benchmark provides empirical evidence of reasoning capabilities, we also assess the quality of the generated text. GPT-4o scores each model's responses from 1-5 based on four key metrics: helpfulness, relevance, accuracy, and level of detail. These evaluations are benchmarked against ground truth responses in the dataset. Each evaluation is performed in triplicate to enhance robustness.

\begin{table}[h]
  \scriptsize
  \setlength{\tabcolsep}{3pt}      
  \centering
  \begin{tabular}{@{}lrrrrr@{}}
    \toprule
    \multicolumn{6}{c}{\bfseries TS Instruct QA}                     \\
    \cmidrule(l){1-6}
    Model        & Overall            & Helpfulness        & Relevance           & Accuracy           & Score \\
    \midrule
    LLama-3.1-8B & \underline{3.99±1.49} & 3.87±1.57          & 3.95±1.60           & 3.94±1.62          & 8     \\
    TS           & 3.33±1.60            & 3.36±1.73          & 3.42±1.77           & 3.38±1.81          & 0     \\
    PreTS        & 3.72±1.49            & 3.74±1.62          & 3.82±1.66           & 3.80±1.69          & 1     \\
    OrcaTS       & 3.92±1.28            & \underline{3.97±1.44} & \textbf{4.16±1.47}  & \textbf{4.14±1.49} & 9     \\
    PreOrcaTS    & \textbf{4.04±1.33}   & \textbf{4.05±1.44}   & \underline{4.13±1.46} & \underline{4.12±1.48} & 12    \\    
    \midrule
    \multicolumn{6}{c}{\bfseries TS Math Analysis}                   \\
    \cmidrule(l){1-6}
    Model        & Overall            & Helpfulness        & Relevance           & Accuracy           & Score \\
    \midrule
    LLama-3.1-8B & 3.09±1.11            & 2.97±1.13          & 2.97±1.16           & 2.83±1.16          & 0     \\
    TS           & 3.66±1.02            & 3.75±1.09          & 3.87±1.08           & 3.42±1.27          & 0     \\
    PreTS        & 3.96±0.75            & \underline{4.05±0.85} & \underline{4.16±0.83} & 3.72±0.98          & 7     \\
    OrcaTS       & \textbf{4.01±0.78}   & \textbf{4.08±0.89}   & \textbf{4.20±0.84}  & \textbf{3.79±1.03} & 14    \\
    PreOrcaTS    & \underline{3.99±0.88} & 4.05±0.97          & 4.16±0.97           & 3.76±1.07          & 11    \\
    \bottomrule
  \end{tabular}
    \caption{Quantitative ablation of Chat-TS variants.  
    Points: 1st=3, 2nd=2, 3rd=1.}
  \label{tab:qa_math_quant}
\end{table}
We conducted this analysis on 100 samples from the TS-Instruct QA Gold benchmark and an additional 100 samples focusing on decision-making and mathematical reasoning tasks. These additional samples were generated separately from the TS-Instruct training dataset, allowing us to test model performance in more complex scenarios outside their training datasets. Results are summarized in Table~\ref{tab:qa_math_quant}. Across both TS Instruct QA Gold and mathematical reasoning tasks, PreOrcaTS and OrcaTS demonstrate superior performance, achieving the highest cumulative score. These results validate the effectiveness of interleaving text and time-series data, not only for improving instruction-following capabilities but also for enhancing the quality of reasoning and explanation generation. 

\subsubsection{Encoding Ablation Analysis}
To compare how our encoding method (discrete tokenization) and training data impact the model performance we produce several ablations shown in Table \ref{tab:matched_baselines_onecol}. 

\begin{table}[h]
\centering
\scriptsize
\setlength{\tabcolsep}{3.5pt}
\renewcommand{\arraystretch}{1.05}
\resizebox{\columnwidth}{!}{%
\begin{tabular}{l c l r r}
\toprule
\textbf{Model} & \textbf{TS Enc.} & \textbf{Training Data} & \textbf{QA Gold (\%)} & \textbf{MCQ2TS (\%)} \\
\midrule
Llama3.1-8B                 & Text        & N/A                  & 54.2 & 36.5 \\
Llama3.1-8B                 & Text        & OpenOrca+TS-Instruct  & 63.0 & 50.8 \\
Llama+MLP                   & MLP         & OpenOrca+TS-Instruct  & 67.5 & 43.1 \\
Llama+Transformer           & Transformer & OpenOrca+TS-Instruct  & 65.9 & 45.0 \\
Chat-TS                     & Tokens      & OpenOrca+TS-Instruct  & 69.4 & 56.4 \\
\bottomrule
\end{tabular}%
}
\caption{\textbf{Fair baselines under matched training/evaluation.} To isolate modeling choices, we train patch-based encoder baselines and a text-encoding baseline using the same base model (Llama3.1-8B), training data (OpenOrca+TS-Instruct), hyperparameters, and evaluation settings. Scores use a single fixed prompt that asks models to respond in the required answer format.}
\label{tab:matched_baselines_onecol}
\vspace{-2mm}
\end{table}

First we show the base Llama3.1-8B model trained using the text encoding method on the TS-Instruct dataset. We can see that while it does improve the quality of the model, the performance falls short of the the discrete (Chat-TS) model trained on the same data but with our encoding method. Finally, we compare our model to two models that encode the data using two trained encoders (MLP and Transformer) which are then projected to the transformer dimension and concatenated with the text embeddings. We can see that the discrete encoding method of Chat-TS outperforms both methods on both benchmarks.

\section{Conclusions}
We investigate the development of LLMs that are capable of time-series reasoning and natural language generation.
Our research provides several key contributions: the TS Instruct Dataset for instruction-tuning time-series models, the human-annotated TS Instruct QA Gold Benchmark for evaluating time-series reasoning, and a detailed quantitative probing set that assesses the quality of model responses, beyond simple accuracy. We establish a strong, state-of-the-art baseline and demonstrate that the proposed framework outperforms existing benchmarks in different categories, while preserving the natural language capabilities of the underlying LLMs. Our work lays a base for future advancements in multimodal reasoning, combining both text and time-series modalities in a unified model. 

\FloatBarrier

\section*{Limitations}
While our Chat-TS models demonstrate strong time-series reasoning capabilities, there are many avenues for future work to build on our framework. Accurate time-series generation has yet to be achieved, as our models struggle to reliably forecast time-series data. Addressing this limitation could enable broader applications such as financial or weather forecasting.
Time-series classification is another area requiring improvement. The models exhibit weak zero-shot performance on this task, often guessing or repeating class predictions. A few potential solutions to this issue is to design classification datasets for multi-modal LLMs which focus on few-shot learning examples may help the model to reason in a zero-shot classification setting. In this case training the model on a larger context window with more compute could allow the model generalize to these scenarios. 

Finally, scaling up this framework both in terms of data volume and model size would almost certainly lead to improved performance~\citep{hoffmann2022trainingcomputeoptimallargelanguage}. We chose for our experiments the LLama-3.1-8B model which has strong performance for the model size, making our research accessible to the broader community.

\section*{Acknowledgments}
We would like to acknowledge the support and funding from Ingenuity Labs Research Institute at Queen's University and Ingenuity Labs Seed Funding. This research was enabled in part by support provided by Calcul Québec and the Digital Research Alliance of Canada. This Research is partially supported by NSERC Discovery Grants.

\bibliography{Chat-TS}

\appendix
\section{Assets}
\label{appendix:assets}
We release relevant assets to the public community below.

\subsection*{Data}
{\raggedright
\noindent{TS Instruct Dataset:}\par
\url{https://huggingface.co/datasets/PaulQ1/TS_Instruct_Reasoning_V1}\par\smallskip

\noindent{TS Instruct QA Gold Benchmark:}\par
\url{https://huggingface.co/datasets/PaulQ1/TS_Instruct_QA_Gold_v2}\par\smallskip

\noindent{TS Instruct Quantitative Evaluation Benchmark:}\par
\url{https://huggingface.co/datasets/PaulQ1/Chat-TS-Quant-Eval}\par
}

\subsection*{Models}
{\raggedright
\noindent The trained Chat-TS model is publicly available on Hugging Face:\par
\url{https://huggingface.co/PaulQ1/Chat_TS}\par
}

\subsection*{Code}
{\raggedright
\noindent The full training, evaluation, and data processing codebase is available on GitHub:\par
\url{https://github.com/quinlanp/Chat-TS-Multi-Modal-Reasoning}\par
}

\section{Related Work}
\label{app:related}

Here we provide an exapanded related works section which contains all of the information from the main body of the paper and additionally talks about the development of foundation models in the time-series domain and how this work differs in its approach to pure time-series modeling methods.

\paragraph{Foundation Models in Time-Series Analysis.} There have been many works which focus building large artificial intelligence models for analysing time-series. These models are often trained on a wide range of time-series and use self-supervised training objectives such as series reconstruction and prediction~\citep{ansari2024chronoslearninglanguagetime,woo2024unifiedtraininguniversaltime,ansari2024chronoslearninglanguagetime}. While these models have been trained on a wide range of different time-series and may act as a core model for different time-series applications~\cite{woo2024unifiedtraininguniversaltime}, they cannot reason over both nautral language and time-series tasks. This makes them generally unable to process text and perform text based tasks (analysis and generation) of multi-modal time-series data. This makes them generally unsuitable for time-series reasoning as defined in this work.

\paragraph{LLMs as Core Models for Time-Series.} Many recent works treat large language models as a backbone for forecasting or classification by repurposing their pretrained parameters, but at the cost of disabling their full text‐generation and reasoning abilities. For instance, GPT4TS~\citep{zhou2023fitsallpowergeneraltime} and LLM4TS~\citep{chang2024llm4tsaligningpretrainedllms} embed series in model weights; TEMPO~\citep{cao2024tempopromptbasedgenerativepretrained}, and TEST~\citep{sun2024testtextprototypealigned} insert series into prompts; and InstructTime~\citep{cheng2024advancingtimeseriesclassification} fine-tunes for classification, without preserving general multimodal reasoning.
LLMTime~\citep{gruver2024largelanguagemodelszeroshot} instead encodes time-series as text, enabling zero-shot forecasting without task-specific fine-tuning and retaining the model’s text-based structure. 

\paragraph{Time-Series Reasoning with LLMs.} A few recent efforts have begun to explore time-series reasoning rather than pure forecasting. ChatTime~\citep{chattime} and TimeMQA~\citep{timemqa} build generated datasets which test time-series alignment (eg. identifying trends) with LLMs but train and test on subsets from the same global datasets which limit comparability. Chow et al.~\citep{chow2024timeseriesreasoningllms} augment an LLM with a learned encoder for raw series inputs; however, neither their code nor the assessment of the model’s natural-language capabilities has been released. In contrast, Chat-TS not only preserves end-to-end text input and generation, but shows strong generalization on our TS-Instruct QA Gold dataset. This dataset focuses on real-world reasoning scenarios, despite only training on simple conversational samples. We publish both model and dataset benchmarks. A concurrent work \citep{xie2025chattsaligningtimeseries} trains LLMs for time-series reasoning. This work uses a trained encoder and embedding concatenation to combine time-series representations. We believe our methods are orthogonal to~\citep{xie2025chattsaligningtimeseries}. For details, please refer to Appendix F.

\paragraph{Time-Series Training and Evaluation.} Standard repositories (Monash~\citep{godahewa2021monash}, UCR~\citep{Middlehurst_2024}, LOTSA~\citep{woo2024unifiedtraininguniversaltime}) offer forecasting and classification benchmarks but lack textual context for multimodal reasoning. Merrill et al.~\citep{merrill2024languagemodelsstrugglezeroshot} evaluate LLM reasoning on synthetic series and LLM‐generated prompts, highlighting performance gaps but without providing training improvement strategies. We address this gap by releasing (1) a large-scale conversational training corpus of human-LLM exchanges grounded in real-world series metadata, and (2) TS-Instruct QA Gold, a human-validated question–answer set over real series with visual context. Merrill also contributed a single-time-series question and answer (QA) based dataset using synthetic dataset. Unfortunately they deemed it not appropriate for LLM evaluation as existing text models had near perfect performance on these benchmarks. In contrast we contribute a challenging QA benchmark which consists of real-world time-series and utilize human evaluators to verify the accuracy of each question.

\paragraph{Discrete Tokenization.} Existing multimodal tokenizers (e.g., VQ-VAE~\citep{oord2018neuraldiscreterepresentationlearning} or channel-wise patching) optimize for compression on very long sequences, often degrading reconstruction fidelity. We introduce a lightweight discrete tokenizer aimed at maximizing accuracy on sub-1,000-point series. While Chronos~\citep{ansari2024chronoslearninglanguagetime} uses similar discrete tokenization in LLM-adjacent models, it does not maintain full LLM text-generation or reasoning capabilities and focuses on time-series forecasting.

\section{Dataset Generation}
\label{APPENDIX: Dataset Generation}
Our training data is organized into three categories: pure time-series data used in pretraining, pure natural language data, and joint time-series and natural language data.
\subsection{Time-Series Data}
\label{subsec:time-seriesdata}
While there are many collections of time-series data across various domains, we utilize the LOTSA~\citep{woo2024unifiedtraininguniversaltime} dataset archive, which is a large compilation spanning multiple domains. Since this dataset is used to pre-train the embeddings and language modeling head for time-series data, our goal is to ensure diversity not only in modalities but also in the physical characteristics of the data, such as sequence length and channel dimensions.
The dataset generation process is described below:

Given a time-series dataset \( \{\bm{X}_i\}_{i=1}^{N} \), where each \( \bm{X}_i \in \mathbb{R}^{T_i \times C} \) represents a time-series with \( T_i \) timesteps and \( C \) channels, we generate time-series samples \( \{\bm{S}_{ij}\}_{j=1}^{n_i} \) using a sliding window approach. A fixed window size \( W \) and stride \( D \) are applied to extract subsequences \( \bm{S}_{ij} \in \mathbb{R}^{w_{ij} \times c_{ij}} \), where \( w_{ij} \) is the length of the \( j \)-th segment and \( c_{ij} \) is the number of channels selected from \( C \). These parameters ultimately control the number of samples parsed from each dataset.

The start index for each subsequence is \( t_{ij} = jD \) for \( j = 0, 1, \dots, \left\lfloor \frac{T_i - W}{D} \right\rfloor \), and the segment length \( w_{ij} \) is determined as \( w_{ij} = \min(W, \text{rand}(m, \min(M, T_i - t_{ij}))) \), where \( m \) and \( M \) represent the minimum and maximum allowable segment lengths. The dimensionality \( c_{ij} \) is selected randomly such that \( 1 \leq c_{ij} \leq C \), and the subsequence \( \bm{S}_{ij} \) is extracted from the original sequence. If the total number of elements \( | \bm{S}_{ij} | = w_{ij} \times c_{ij} \) exceeds a specified maximum, \( c_{ij} \) is reduced iteratively. This process is repeated for all time-series in the dataset, yielding a set of samples \( \{\bm{S}_{ij}\} \) constrained by a maximum sample size, with a wide range of physical properties.

\subsection{Natural Language Data}
For the natural language data, we use the Open-Orca ~\citep{OpenOrca} dataset, which has been successfully used to train high-quality instruction models and contains a large number of samples. This data is combined with the time-series instruction tuning data to regularize training and preserve the model's performance on natural language tasks. To reduce overall training time, we use 100,000 samples from this dataset during full-parameter instruction tuning of the model.

\subsection{Time-Series and Natural Language Data}
\label{sec:ts_nlp_databuilding}
Currently, there are no joint datasets combining both time-series and natural language data. To train our multi-modal model, we generate synthetic conversations between a user and an assistant using GPT4o-mini~\citep{openai2024gpt4technicalreport}, incorporating both modalities.  

The process begins by sampling $n$ time-series from a dataset \( \mathbf{D} = \{ \mathbf{x}_1, \mathbf{x}_2, \dots, \mathbf{x}_N \} \), where $n = \min(\text{len}(\mathbf{D}), N_s)$ samples are selected, $N_s$ being the maximum samples per dataset. For each time-series \( \mathbf{x}_i \), the corresponding target data \( \mathbf{y}_i \in \mathbb{R}^{T \times C} \), where \( T \) is the length and \( C \) the number of channels, is also extracted. A random number of channels, \( C' \in [1, C] \), and a time-series length \( T \) (proportional to the full length, constrained by \( L_{\max} \)) are selected. 

The series lengths are computed as:
\[
T = \max\left( 1, \min\left( \left\lfloor L_{\text{max}} \cdot U(\alpha_{\min}, \alpha_{\max}) \right\rfloor, L_{\max} \right) \right),
\]
where $U(a, b)$ is a uniform distribution, and \( \alpha_{\min} \) and \( \alpha_{\max} \) are the minimum and maximum percentages of the sequence used. 

To create conversations, we use a system prompt, an image of the time-series, and pass the length and number of channels of the time-series. The system prompt provides explicit instructions for generating the conversation, and the time-series dimensions (length, channels) are also provided to ensure the model uses them during the conversation. Since GPT4o-mini cannot analyze the time-series directly, we attach an image to guide the generated conversation based on the trends and patterns within the time-series. 

We design four types of conversations: (1) \textbf{General Reasoning}: These focus on descriptive discussions of time series data, highlighting key aspects and trends; (2) \textbf{Classification}: This involves categorizing the time series based on user-provided labels; (3) \textbf{Decision-Making}: The model evaluates hypothetical scenarios posed by the user, drawing conclusions from the time series data; and (4) \textbf{Mathematical Reasoning}: This covers in-depth mathematical analysis of time series. Full details and examples can be found in Appendix \ref{Appendix_Time-series_and_Text}, \ref{APPENDIX: System_Prompts} and \ref{Sec:Extended_Case_Study}. General reasoning, decision-making and mathematical reasoning time-series are from the LOTSA dataset compilation ~\citep{woo2024unifiedtraininguniversaltime}, the classification samples and labels are from the Time-Series Classification Archive~\citep{Middlehurst_2024}.
\label{Appendix_Time-series_and_Text}

\subsubsection{Dataset Overview}
\begin{figure}[h]
    \centering
    \includegraphics[width=0.5\textwidth]{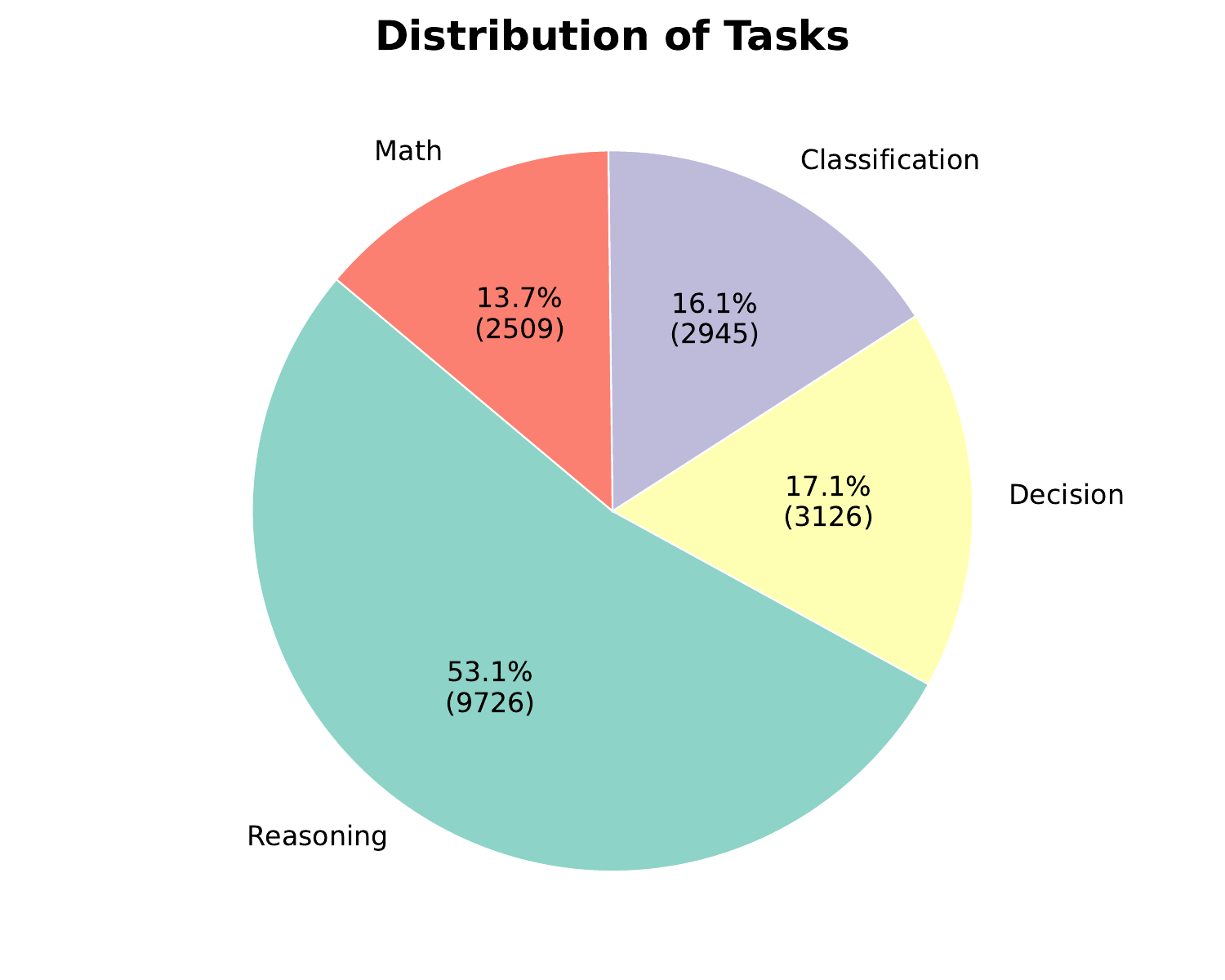}
    \caption{Breakdown of sample types in the TS Instruct dataset. In total we have 18306 total samples in our dataset.}
    \label{fig:task_distribution}
\end{figure}

Our TS Instruct dataset contains 18306 samples generated using a combination of time-series properties, visual representations and detailed instructions to elucidate a diverse set of instructing samples for time-series instruction tuning. A breakdown of the dataset composition is shown in Figure \ref{fig:task_distribution}. We clean the dataset by removing samples with incorrect formatting, repeated conversation roles, lack of $[user-input]$ placeholder for the correct placement of time-series within the conversation or any incorrect universally unique identifier (UUID) linked to each sample to ensure the text and time-series for each sample can be easily linked together at tokenization time.

\subsubsection{Token Distribution}
\begin{figure*}[h]
    \centering
    \includegraphics[width=0.75\textwidth]{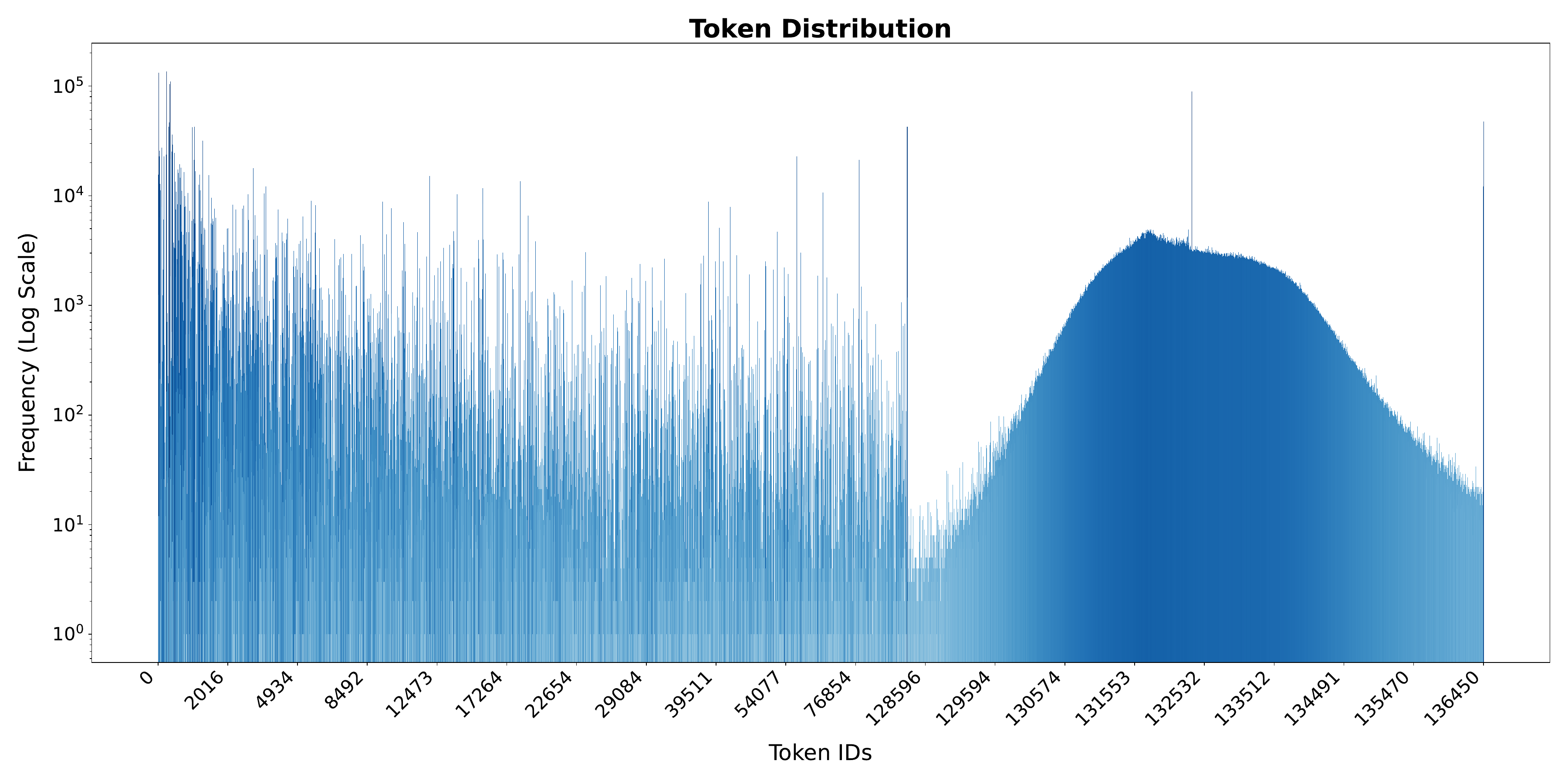}
    \caption{This plot shows the frequency distribution of token IDs in log scale. The x-axis represents token IDs, and the y-axis represents the frequency of occurrences on a logarithmic scale.}
    \label{fig:token_distribution_all}
\end{figure*}

We tokenize our TS Instruct dataset using the discrete tokenizer as described. We add these new tokens to the model vocabulary and can measure the token representation across the dataset for time-series. The data visualized in Figure \ref{fig:token_distribution_all} shows that the time-series data follows a normal distribution. Notably since our tokenizer simply discretizes the time-series there is minimal distortion between the tokenized and non-tokenized distributions. Interestingly the text tokens within the vocabulary are relatively uniformly distributed. This reinforces the quality of our generated dataset since the text instructions and responses are diverse.

\subsection{TS Instruct Dataset Example}
See the TS Instruct dataset example below in Table \ref{ts_instruction_tuning_examples}.
\begin{table*}[h!]
\centering
\renewcommand{\arraystretch}{0.8}
\begin{tabular}{p{0.6\textwidth} p{0.34\textwidth}}
    \hline
    \multicolumn{2}{l}{\textbf{Instruction-Tuning Dataset Example 1}} \\
    \hline

    \parbox[t]{\hsize}{\vspace{0pt}%
        \textbf{User:} I've been monitoring the power consumption in my house. Here's the data I've collected so far.\\[2mm]
        \textbf{Assistant:} Interesting! The power consumption data showcases quite a bit of variability. There are frequent fluctuations with notable peaks toward the end. This suggests occasional high usage, possibly from appliances or events that significantly increase energy draw. The troughs indicate periods of low consumption, hinting at more consistent, lower-energy activities during these times.
        \\
    }
    &
    \parbox[t]{\hsize}{\vspace{0pt}%
        \centering
        \includegraphics[width=\linewidth, height=0.5\textheight, keepaspectratio]{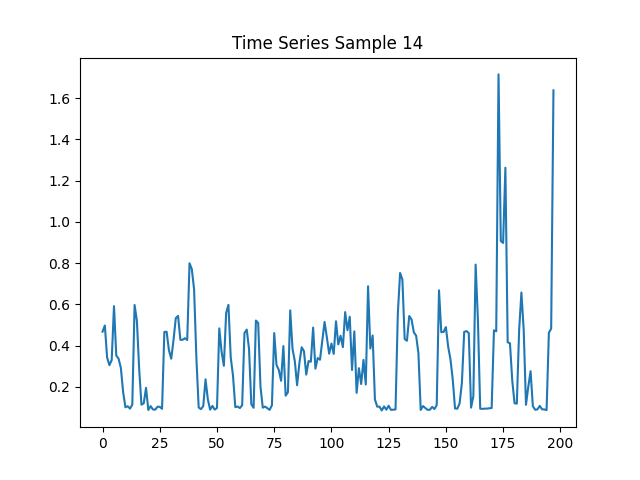}
    }
    \\
    \hline

    \multicolumn{2}{l}{\textbf{Instruction-Tuning Dataset Example 2}} \\
    \hline

    \parbox[t]{\hsize}{\vspace{0pt}%
        \textbf{User:} I've been monitoring some data recently and here's what it looks like.\\[2mm]
        \textbf{Assistant:} It looks like there are some significant fluctuations in the data. Initially, there's a tall spike early on, followed by a series of smaller oscillations. Around the midpoint, the activity becomes much more pronounced with higher peaks and greater variability. Toward the end, there’s a noticeable decrease in the fluctuations, returning to a more subdued state. This could indicate a period of increased activity or disturbances followed by stabilization.
        \\
    }
    &
    \parbox[t]{\hsize}{\vspace{0pt}%
        \centering
        \includegraphics[width=\linewidth, height=0.5\textheight, keepaspectratio]{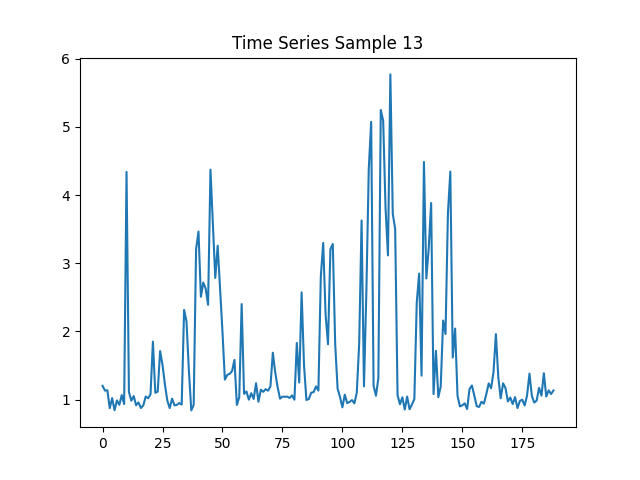}
    }
    \\
    \hline

    \multicolumn{2}{l}{\textbf{Instruction-Tuning Dataset Example 3}} \\
    \hline

    \parbox[t]{\hsize}{\vspace{0pt}%
        \textbf{User:} I've been analyzing some sensor readings over a short period. Here's the time-series data:\\[2mm]
        \textbf{Assistant:} This time-series showcases a significant spike at the 10th point, which suggests a moment of intense activity or a sudden event impacting the readings. It's interesting that the data returns almost immediately to a low level, indicating that the event was quite brief. The stable low values surrounding this peak might imply regular monitoring with occasional disruptions. Any idea what caused that spike?
        \\
    }
    &
    \parbox[t]{\hsize}{\vspace{0pt}%
        \centering
        \includegraphics[width=\linewidth, height=0.5\textheight, keepaspectratio]{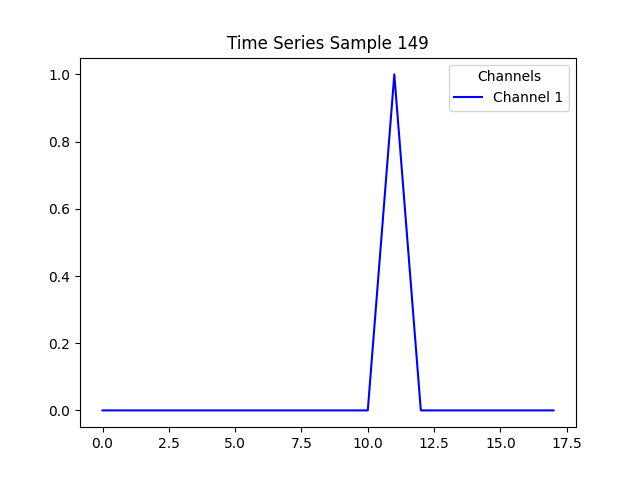}
    }
    \\
    \hline

\end{tabular}
\vspace{2mm}

\caption{Examples from the TS Instruct Dataset.}
\label{ts_instruction_tuning_examples}
\end{table*}

\subsection{TS Instruct QA Gold Dataset Example}
Example TS Instruct QA Gold example is shown below in Table \ref{ts_qa_example}.
\begin{table*}[h!]
\centering
\renewcommand{\arraystretch}{1.5}
\begin{tabular}{p{0.95\textwidth}} 
    \hline
    \textbf{Multi-Modal QA Example} \\ 
    \hline

    \parbox{0.95\textwidth}{ 
        \vspace{2mm}
        \textbf{Question:} The dataset represents fluctuations in energy consumption over a specified period. Based on this, what could be the most likely explanation for the observed trends?
        \centering
        \includegraphics[width=0.4\textwidth]{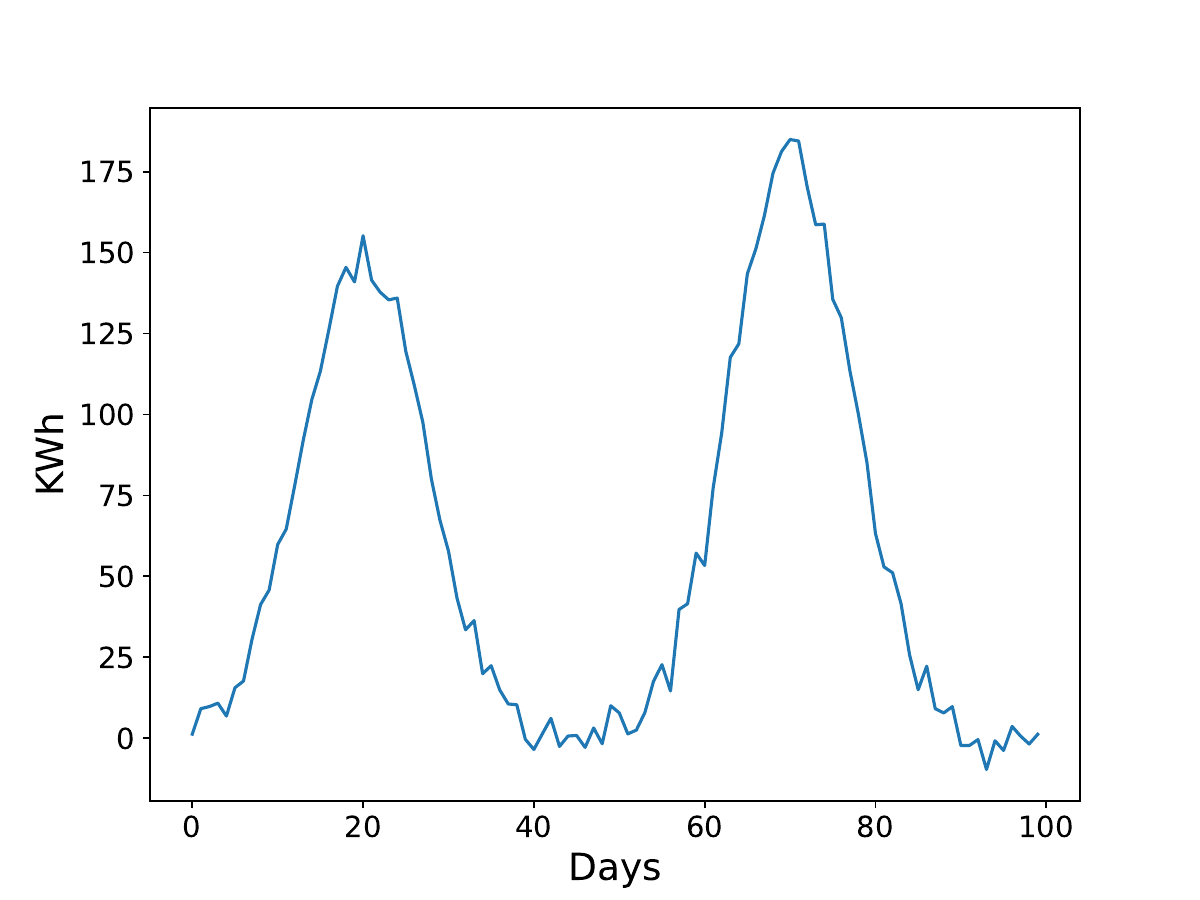} \\
    } \\
    A: Seasonal changes in energy demand.  \\ 
    B: An error in data collection.  \\ 
    C: A stationary consumption pattern across the entire period.  \\ 
    D: A sudden decrease in the population. \\ 
    \hline

    \textbf{Answer:} The correct answer is A: Seasonal changes in energy demand. \\ 
    \textit{(Optional explanation):} The significant peaks and troughs in the data suggest a cyclical trend, commonly associated with seasonal changes in energy consumption—for example, higher demand during summer and winter months due to heating and cooling needs. The other options do not align with the observed fluctuations in the data. \\
    \hline

    \textbf{LLama 3.1-8B Model Answer:} A: Seasonal changes in energy demand. \\ 
    \textit{Explanation:} The given dataset represents fluctuations in energy consumption over a specified period. The observed trends in the data suggest a cyclical pattern, with peaks and troughs occurring at regular intervals. This is indicative of seasonal changes in energy demand, which can be influenced by factors such as temperature, humidity, and human activity. For instance, energy consumption tends to be higher during winter months when heating is required and lower during summer months when cooling is not as necessary. \textcolor{professionalred}{The data also shows a general upward trend, which could be due to an increase in population, economic growth, or other factors that contribute to increased energy demand.} \\
    \hline

    \textbf{Chat-TS Model Answer:} A: Seasonal changes in energy demand. \\ 
    \textit{Explanation:} The most likely explanation for the observed trends in the dataset is seasonal changes in energy demand. This is because energy consumption often varies with the seasons, with higher consumption during colder months and lower consumption during warmer months. The fluctuations in the dataset could reflect these seasonal patterns, indicating that energy usage is influenced by external factors such as temperature and weather conditions. \\
    \hline
\end{tabular}
\vspace{2mm}
\caption{Example question and responses from our multiple-choice QA benchmark, including the correct answer, the LLama 3.1-8B model (state-of-the-art baseline), and the PreOrcaTS variation of the Chat-TS models. We also highlighted incorrect portions of the explanation in \textcolor{professionalred}{red} showing the importance of both testing model accuracy and performing quantitative evaluation on model responses.}
\label{ts_qa_example}
\end{table*}

\section{Training Details}
\label{APPENDIX:Training_Details}
Here we highlight some important hyper parameters and training details to help make our work more reproducible. We also provide all of the code for training our models.

\begin{table*}[htbp]
\centering
\caption{Training Configurations for Tokenizer and LLM}
\label{tab:training_configurations}
\begin{tabularx}{0.9\textwidth}{l l >{\centering\arraybackslash}X >{\centering\arraybackslash}X}
\hline
\multicolumn{2}{c}{\textbf{Tokenizer Configuration}} & \multicolumn{2}{c}{\textbf{LLM Configuration}} \\
\hline
\textbf{Parameter} & \textbf{Value} & \textbf{Parameter} & \textbf{Value} \\
\hline
Sequence Length & 1024 & Context Length & 2048 \\
Hidden Dimension & 64 & Batch Size (Training) & 4 \\
Number of Tokens & 1024 & Gradient Accumulation Steps & 16 \\
Codebook Size & 8192 & Precision & bf16 \\
\hline
\multicolumn{2}{c}{\textbf{Tokenizer Training Configuration}} & \multicolumn{2}{c}{\textbf{LLM Training Configuration}} \\
\hline
\textbf{Parameter} & \textbf{Value} & \textbf{Parameter} & \textbf{Instruction Tuning \,|\, Pretraining} \\
\hline
Window Size & 256 & Epochs & 1 / 1 \\
Sliding Window Step Size & 32 & Learning Rate & 2e-5 / 2e-3 \\
Batch Size & 256 & Optimizer & Adam / Adam \\
Learning Rate & 0.001 & FSDP & Full Shard / Full Shard \\
Epochs & 3 & Gradient Clipping & 1.0 / 1.0 \\
Precision & bf16 & Seed & 42 / 42 \\
Seed & 42 & Warmup Steps & 10\% / 10\% \\
Gradient Clipping & 1.0 & & \\
Warmup Steps & 1000 & & \\
\hline
\end{tabularx}
\end{table*}

All LLM training was performed on a cluster with 4 NVIDIA A100 GPUs (40GB each). All tokenizer training was performed on a single NVIDIA A100 GPU (40GB). Tokenizer training took roughly ~(0.5-1) hours, Pretraining ~4 hours and Instruction tuning ~3.5 hours. 

\section{System Prompts}
\label{APPENDIX: System_Prompts}
We utilize system prompts to as a method for controlling how the LLM generates our multi-modal datasets. Our system prompts contain three componenets 1) Guidelines - These are used to control the type of conversation 2) Formatting - We want the conversation to have specific formatting to enable the injection of roles and time-series 3) Examples - We found that adding an example to the system prompt greatly increased the success rate and quality during dataset generation.

These prompts are \textbf{combined} with properties of the time-series such as length and number of channels, information about the type of time-series and an image of the time-series to ensure that the conversation is accurate.

\subsection{TS Instruct Training Data Generation}
Below in Figure \ref{fig:reasoning_sys_prompt} is an example of the system prompt used for generating reasoning samples. The goal of this conversation is to create conversations based around events and properties of the time-series. This data helps learn trends and properties in an instruction format. We have found that providing a few different examples at the end of the system prompt not only helps with the formatting of the conversations but also in generating diverse samples with differing tones. 

\begin{figure*}
\refstepcounter{figure}\label{fig:reasoning_sys_prompt} 
\addcontentsline{lof}{figure}{System prompt example}
\begin{tcolorbox}[colback=lightgreen, colframe=green!50!black, title=Generate a simple conversation focused on reasoning about time-series data.]
\textbf{Guidelines}:
\begin{enumerate}
    \item The user initiates the conversation by providing a time-series using the \textbf{[user-input]} placeholder.
    \item The AI internally analyzes the time-series and provides accurate insights, reasoning, and observations based on the data.
    \item The AI provides an accurate description of the time-series based on the image without making direct references to the image.
    \item Ensure each conversation is unique and tailored to the specifics of the provided time-series.
\end{enumerate}

\textbf{Formatting}:
\begin{itemize}
    \item Use \textbf{User:} and \textbf{AI:} roles.
    \item Each conversation starts with \textbf{User:} and the AI's response follows.
    \item The \textbf{[user-input]} placeholder should be used once in the first user message.
    \item Keep the conversation length to 2 turns (User input and AI response).
    \item Utilize a random conversation style (\textit{formal, assertive, casual, goofy}, etc.).
    \item Reference a specific component in the time-series.
\end{itemize}

\vspace{0.5cm}


\begin{tcolorbox}[colback=lightgreen!80, colframe=green!50!black, title=Example 1: Analytical and Detailed]
\textbf{User}: I have been tracking the daily temperatures in my city for the past month. Here is the time-series data: \textbf{[user-input]}\\
\textbf{AI}: The time-series shows a clear fluctuation between the first 12 days of the month with a pronounced dip in temperatures during the second week. This is followed by a steady upward trend in the latter half towards the 30th day. The oscillations could suggest changes in local weather patterns or an isolated cold event.
\end{tcolorbox}

\vspace{0.3cm}

\begin{tcolorbox}[colback=lightgreen!80, colframe=green!50!black, title=Example 2: Casual and Conversational]
\textbf{User}: I’ve been keeping tabs on my energy usage over the last year. Here’s what the data looks like: \textbf{[user-input]}\\
\textbf{AI}: Oh, I see some pretty big spikes in energy use, particularly around input 143, especially during the colder months. Looks like your heater's been working overtime! The dips, on the other hand, at points 18, 27, and 64, seem to line up with warmer periods—saves you some energy, huh?
\end{tcolorbox}

\vspace{0.3cm}

\begin{tcolorbox}[colback=lightgreen!80, colframe=green!50!black, title=Example 3: Curious and Thoughtful]
\textbf{User}: Here are my sleep patterns for the past few months. \textbf{[user-input]}\\
\textbf{AI}: Hm, it’s interesting how your sleep cycles fluctuate. You can see a more regular pattern toward the beginning until the 3rd month, but around September, the data suggests some interruptions—maybe related to stress or changes in routine?
\end{tcolorbox}

\end{tcolorbox}
\end{figure*}

We also provide an example in Figure \ref{fig:Math_Sys_Prompt} of the system prompt used to generate the samples for mathematical reasoning. This is a truncated version in order to meet the page requirements, please refer to our code for the full prompt. The full prompt contains 5 total examples of interesting math related analysis of time-series data.

\begin{figure*}
\begin{tcolorbox}[colback=softpurple, colframe=purple!50!black, title=Task: Generate a conversation where the AI solves a problem or performs calculations based on real-time-series data provided by the user.]
\refstepcounter{figure}\label{fig:Math_Sys_Prompt} 
\addcontentsline{lof}{figure}{Math Sys Prompt Example}
\textbf{Guidelines}:
\begin{enumerate}
    \item The AI must use \textbf{real numbers} from the time-series data provided by the user and perform calculations based on that data.
    \item The user will provide context for a problem or question requiring a mathematical solution (e.g., calculating slopes, rate of change, area under the curve, etc.).
    \item The AI must perform calculations on the \textbf{[user-input]} time-series data and show the work step-by-step. \textbf{No hypothetical data} or generic placeholders should be used.
    \item The AI must refer to specific trends, patterns, and numerical values from the provided data.
    \item The AI’s response should include the relevant formula and apply it using real data, explaining the calculation process clearly and accurately.
    \item The conversation should include mathematical diversity beyond averages, including tasks like but not limited to computing rate of change, slope, area under the curve, etc.
    \item The conversation should include two turns:
    \begin{itemize}
        \item The user’s message, providing context and the time-series data.
        \item The AI’s response, including the solution and a clear explanation of the calculations.
    \end{itemize}
\end{enumerate}

\textbf{Formatting}:
\begin{itemize}
    \item Use \textbf{User:} and \textbf{AI:} roles to define the conversation.
    \item The \textbf{[user-input]} placeholder should be used once in the user's message.
    \item Ensure the AI calculates directly using the provided data and presents a complete, step-by-step solution.
\end{itemize}

\vspace{0.5cm}

\textbf{Example conversation (as part of the system prompt)}:

\begin{tcolorbox}[colback=softpurple!80, colframe=purple!50!black, title=Example 1: Focus on Slope (casual tone)]
\textbf{User}: I've been tracking sales data for the past few quarters, and I need to know the overall trend. Can you calculate the slope of the sales data for me? \textbf{[user-input]}\\
\textbf{AI}: Sure! To calculate the slope of the sales data, we’ll use the formula for the slope:

   \[
   m = \frac{y_2 - y_1}{x_2 - x_1}
   \]

Based on the provided data points over four quarters, we have the following sales (in units):

   \[
   Q1 = 120, Q2 = 150, Q3 = 180, Q4 = 210
   \]

Now, calculating the slope between Q1 and Q4:

   \[
   m = \frac{210 - 120}{4 - 1} = \frac{90}{3} = 30
   \]

So, the sales are increasing by 30 units per quarter, indicating a positive upward trend.
\end{tcolorbox}

\vspace{0.3cm}

\end{tcolorbox}
\end{figure*}
\subsection{TS Instruct QA Gold Benchmark}
Within the QA benchmark we want the user to create a realistic scenario which require reasoning over the time-series to answer the question. These generally have to do with analyzing the trend and picking a the scenario which matches that trend. The prompt is shown below in Figure \ref{fig: QA_Sys_Prompt}.

\begin{figure*}
\begin{tcolorbox}[colback=lightblue, colframe=blue!50!black, title=Task: Generate a simple multiple-choice conversation regarding complex reasoning with time-series.]
\refstepcounter{figure}\label{fig: QA_Sys_Prompt} 
\addcontentsline{lof}{figure}{QA System prompt}
\textbf{Guidelines}:
\begin{enumerate}
    \item First pick a random theme. Examples include but are not limited to \textit{energy, economics, environmental, motion, healthcare}, etc.
    
    \item The user will create a backstory with context, without directly describing the trends in the time-series data. Use the \textbf{[user-input]} placeholder for the time-series data that the AI must analyze.
   
    \item The user will provide multiple plausible options for reasoning about the specific time-series data, focusing on correct vs. flawed interpretations of the time-series.

    \item The AI will refer to the provided image as the time-series data source without directly mentioning the image itself. The AI will answer as though it is directly analyzing the time-series data and detecting the trends within it.

    \item Optionally, after choosing the correct option, the AI can explain its reasoning based on the time-series data's specific trends or patterns.
\end{enumerate}

\textbf{Formatting}:
\begin{itemize}
    \item Use \textbf{User:} and \textbf{AI:} roles for the conversation.
    \item Each conversation begins with \textbf{User:} providing the background information and a multiple-choice question, while the time-series trends are contained in the \textbf{[user-input]} placeholder.
    \item The \textbf{[user-input]} placeholder represents the time-series trends, and the AI must analyze the data to identify the trends.
    \item Provide Multiple choice options with each option addressing the specific observations in the time-series. The letter \textbf{D} should be the correct answer.
    \item The \textbf{AI} should follow with the correct answer and optionally provide an explanation directly related to the trends within the time-series data.
\end{itemize}

\vspace{0.5cm}

\begin{tcolorbox}[colback=lightblue!80, colframe=blue!50!black, title=Example Conversation:]
\textbf{User}: The dataset represents temperature variations over the last decade. Based on this, what could be the most likely cause? \textbf{[user-input]}

\begin{itemize}
    \item A: Increased solar activity  
    \item B: Increased industrial emissions  
    \item C: A significant decrease in volcanic activity causing cooling  
    \item D: Ocean currents weakening over time causing cooling
\end{itemize}

\textbf{AI}: The correct answer is \textbf{B: Increased industrial emissions.}

(Optional explanation): The steady upward trend over the past few years is more consistent with human-driven factors such as industrial emissions rather than natural events like solar activity, volcanic changes, or ocean currents.
\end{tcolorbox}
\end{tcolorbox}
\end{figure*}

\subsection{TS Instruct QA Gold Prompts}
\label{APPENDIX:SUBSEC QA_PROMPTS}
In this section, we display the system prompts used to evaluate the models on the TS Instruct dataset. Since the prompt may significantly affect performance, it is essential to test various prompts. Each prompt's primary goal is to ensure that the task is clear and the output format is explicit. The combined prompts are shown below in Figure \ref{fig:qa_prompts_combined}.

\begin{figure*}[h]
    \centering
    \begin{tcolorbox}[colback=blue!5!white, colframe=blue!75!black, boxrule=0.5mm, width=\textwidth, arc=3mm, outer arc=3mm]
    
    \textbf{System Prompt 1:}\\
    You are a highly intelligent assistant designed to analyze time-series data and provide responses based on it. Your role is to help the user by interpreting the data accurately, answering questions clearly, and providing detailed insights. Always ensure that your responses are concise, correct, and grounded in the data you are given.\\
    Your purpose is to assist in decision-making, analysis, and predictions based on time-series data. When asked for multiple-choice answers, always respond in the required \texttt{A:}, \texttt{B:}, \texttt{C:}, or \texttt{D:} format. If a user asks for insights beyond your expertise, politely inform them and suggest alternative actions.
    
    \vspace{5mm} \hrule \vspace{5mm}  
    
    \textbf{System Prompt 2:}\\
    Your purpose is to assist in decision-making, analysis, and predictions based on time-series data. When asked for multiple-choice answers, always respond in the required \texttt{A:}, \texttt{B:}, \texttt{C:}, or \texttt{D:} format.
    
    \vspace{5mm} \hrule \vspace{5mm}  
    
    \textbf{System Prompt 3:}\\
    You are a time-series analyst specializing in decision-making, providing analytical insights on time-series data. When prompted with multiple-choice questions, ensure you adhere to the format: \texttt{A:}, \texttt{B:}, \texttt{C:}, or \texttt{D:}.
    
    \vspace{5mm} \hrule \vspace{5mm}  
    
    \textbf{System Prompt 4:}\\
    Your core purpose is to assist with data-driven decision-making, conduct thorough analysis, and make predictions grounded in time-series data and trend recognition. When presented with multiple-choice questions, always structure your responses in the designated format: \texttt{A:}, \texttt{B:}, \texttt{C:}, or \texttt{D:}, ensuring clarity and alignment with the query requirements.
    
    \vspace{5mm} \hrule \vspace{5mm}  
    
    \textbf{System Prompt 5:}\\
    You are an assistant whose job is:
    \begin{enumerate}
        \item Always respond in the required \texttt{A:}, \texttt{B:}, \texttt{C:}, or \texttt{D:} format.
        \item Provide an explanation.
    \end{enumerate}
    
    \end{tcolorbox}
    \caption{System Prompts used for TS Instruct QA Gold benchmark Evaluation}
    \label{fig:qa_prompts_combined}
\end{figure*}

\section{Text Evaluation Results}
\label{APPENDIX:Text_eval}
We use three different benchmarks in evaluating our models for text analysis. They can be found below:

\begin{itemize}
    \item \textbf{MMLU-Pro (5-shots, multiple-choice):} An enhanced version of the Massive Multitask Language Understanding benchmark, with a focus on reasoning-based questions and expanded choices, providing a more robust and challenging evaluation.
    \item \textbf{BBH (3-shots, multiple-choice):} A collection of 23 challenging tasks from BIG-Bench, focusing on tasks where prior language model evaluations did not outperform average human raters.
    \item \textbf{GPQA (0-shot, multiple-choice):} A graduate-level "Google-proof" question-answering benchmark with difficult questions in biology, physics, and chemistry, designed to test the boundaries of current AI models.
\end{itemize}

Additionally, we present the full results for each model and test in Tables \ref{MMLU_full}, \ref{BBH_full} and \ref{GPQA_full}.

\begin{table*}[htbp]
\centering
\caption{MMLU Pro}
\label{MMLU_full}
\resizebox{\textwidth}{!}{%
\begin{tabular}{l c c c c c c}
Model & LLama 3.1-8B & Orca & OrcaTS & PreOrcaTS & PreTS & TS \\ 
\hline
MMLU Pro & 0.3763 $\pm$0.0044 & 0.3675 $\pm$0.0044 & 0.3566 $\pm$0.0044 & 0.3556 $\pm$0.0044 & 0.3628 $\pm$0.0044 & 0.3603 $\pm$0.0044 \\
\hline
\end{tabular}
}
\end{table*}

\begin{table*}[htbp]
\centering
\caption{Big Bench Hard full results.}
\label{BBH_full}
\resizebox{\textwidth}{!}{%
\begin{tabular}{l c c c c c c}
Subtask & LLama 3.1-8B & Orca & OrcaTS & PreOrcaTS & PreTS & TS \\ 
\hline
boolean\_expressions & 0.8240 $\pm$0.0241 & 0.8120 $\pm$0.0248 & 0.8120 $\pm$0.0248 & 0.8120 $\pm$0.0248 & 0.8320 $\pm$0.0237 & 0.8400 $\pm$0.0232 \\
causal\_judgement & 0.5668 $\pm$0.0363 & 0.6150 $\pm$0.0357 & 0.6471 $\pm$0.0350 & 0.6471 $\pm$0.0350 & 0.6310 $\pm$0.0354 & 0.6096 $\pm$0.0358 \\
date\_understanding & 0.4600 $\pm$0.0316 & 0.4480 $\pm$0.0315 & 0.4280 $\pm$0.0314 & 0.4240 $\pm$0.0313 & 0.4560 $\pm$0.0316 & 0.4440 $\pm$0.0315 \\
disambiguation\_qa & 0.5280 $\pm$0.0316 & 0.6360 $\pm$0.0305 & 0.5640 $\pm$0.0314 & 0.5800 $\pm$0.0313 & 0.6200 $\pm$0.0308 & 0.6280 $\pm$0.0306 \\
formal\_fallacies & 0.5600 $\pm$0.0315 & 0.5400 $\pm$0.0316 & 0.5800 $\pm$0.0313 & 0.5560 $\pm$0.0315 & 0.5520 $\pm$0.0315 & 0.5440 $\pm$0.0316 \\
geometric\_shapes & 0.3640 $\pm$0.0305 & 0.3680 $\pm$0.0306 & 0.3560 $\pm$0.0303 & 0.3280 $\pm$0.0298 & 0.3600 $\pm$0.0304 & 0.3840 $\pm$0.0308 \\
hyperbaton & 0.6280 $\pm$0.0306 & 0.7680 $\pm$0.0268 & 0.7240 $\pm$0.0283 & 0.7400 $\pm$0.0278 & 0.6320 $\pm$0.0306 & 0.6960 $\pm$0.0292 \\
logical\_deduction\_five\_objects & 0.4240 $\pm$0.0313 & 0.4480 $\pm$0.0315 & 0.4160 $\pm$0.0312 & 0.4160 $\pm$0.0312 & 0.4440 $\pm$0.0315 & 0.4480 $\pm$0.0315 \\
logical\_deduction\_seven\_objects & 0.4200 $\pm$0.0313 & 0.4640 $\pm$0.0316 & 0.4320 $\pm$0.0314 & 0.4280 $\pm$0.0314 & 0.4680 $\pm$0.0316 & 0.4560 $\pm$0.0316 \\
logical\_deduction\_three\_objects & 0.6640 $\pm$0.0299 & 0.6400 $\pm$0.0304 & 0.6480 $\pm$0.0303 & 0.6240 $\pm$0.0307 & 0.6600 $\pm$0.0300 & 0.6760 $\pm$0.0297 \\
movie\_recommendation & 0.6480 $\pm$0.0303 & 0.7000 $\pm$0.0290 & 0.6520 $\pm$0.0302 & 0.6720 $\pm$0.0298 & 0.6680 $\pm$0.0298 & 0.6640 $\pm$0.0299 \\
navigate & 0.5480 $\pm$0.0315 & 0.6320 $\pm$0.0306 & 0.6360 $\pm$0.0305 & 0.6400 $\pm$0.0304 & 0.6200 $\pm$0.0308 & 0.5680 $\pm$0.0314 \\
object\_counting & 0.4680 $\pm$0.0316 & 0.5000 $\pm$0.0317 & 0.4640 $\pm$0.0316 & 0.4600 $\pm$0.0316 & 0.4960 $\pm$0.0317 & 0.4760 $\pm$0.0316 \\
penguins\_in\_a\_table & 0.4452 $\pm$0.0413 & 0.5000 $\pm$0.0415 & 0.4932 $\pm$0.0415 & 0.5205 $\pm$0.0415 & 0.4726 $\pm$0.0415 & 0.4795 $\pm$0.0415 \\
reasoning\_about\_colored\_objects & 0.6120 $\pm$0.0309 & 0.6360 $\pm$0.0305 & 0.6240 $\pm$0.0307 & 0.6440 $\pm$0.0303 & 0.6160 $\pm$0.0308 & 0.6000 $\pm$0.0310 \\
ruin\_names & 0.6280 $\pm$0.0306 & 0.7080 $\pm$0.0288 & 0.6440 $\pm$0.0303 & 0.6840 $\pm$0.0295 & 0.7040 $\pm$0.0289 & 0.6320 $\pm$0.0306 \\
salient\_translation\_error\_detection & 0.5120 $\pm$0.0317 & 0.5200 $\pm$0.0317 & 0.4880 $\pm$0.0317 & 0.4760 $\pm$0.0316 & 0.4600 $\pm$0.0316 & 0.4560 $\pm$0.0316 \\
snarks & 0.6798 $\pm$0.0351 & 0.6798 $\pm$0.0351 & 0.7135 $\pm$0.0340 & 0.6629 $\pm$0.0355 & 0.6629 $\pm$0.0355 & 0.7247 $\pm$0.0336 \\
sports\_understanding & 0.6680 $\pm$0.0298 & 0.6720 $\pm$0.0298 & 0.6600 $\pm$0.0300 & 0.6880 $\pm$0.0294 & 0.6560 $\pm$0.0301 & 0.6160 $\pm$0.0308 \\
temporal\_sequences & 0.3480 $\pm$0.0302 & 0.2480 $\pm$0.0274 & 0.1960 $\pm$0.0252 & 0.2200 $\pm$0.0263 & 0.1680 $\pm$0.0237 & 0.1880 $\pm$0.0248 \\
tracking\_shuffled\_objects\_five\_objects & 0.2200 $\pm$0.0263 & 0.1800 $\pm$0.0243 & 0.1680 $\pm$0.0237 & 0.1760 $\pm$0.0241 & 0.2080 $\pm$0.0257 & 0.2320 $\pm$0.0268 \\
tracking\_shuffled\_objects\_seven\_objects & 0.1560 $\pm$0.0230 & 0.1840 $\pm$0.0246 & 0.1920 $\pm$0.0250 & 0.2040 $\pm$0.0255 & 0.1720 $\pm$0.0239 & 0.2000 $\pm$0.0253 \\
tracking\_shuffled\_objects\_three\_objects & 0.3560 $\pm$0.0303 & 0.2880 $\pm$0.0287 & 0.2760 $\pm$0.0283 & 0.2920 $\pm$0.0288 & 0.3280 $\pm$0.0298 & 0.3240 $\pm$0.0297 \\
web\_of\_lies & 0.5320 $\pm$0.0316 & 0.5560 $\pm$0.0315 & 0.4760 $\pm$0.0316 & 0.5000 $\pm$0.0317 & 0.4840 $\pm$0.0317 & 0.4920 $\pm$0.0317 \\
\hline
\end{tabular}
}
\end{table*}

\begin{table*}[htbp]
\centering
\caption{GPQA Full Results}
\label{GPQA_full}
\resizebox{\textwidth}{!}{%
\begin{tabular}{lcccccc}

Task & LLama 3.1-8B & Orca & OrcaTS & PreOrcaTS & PreTS & TS \\ 
\hline
GPQA Diamond & 0.3182 $\pm$0.0332 & 0.3687 $\pm$0.0344 & 0.3889 $\pm$0.0347 & 0.3889 $\pm$0.0347 & 0.3333 $\pm$0.0336 & 0.3586 $\pm$0.0342 \\
GPQA Extended & 0.2985 $\pm$0.0196 & 0.3022 $\pm$0.0197 & 0.3095 $\pm$0.0198 & 0.3132 $\pm$0.0199 & 0.3077 $\pm$0.0198 & 0.3022 $\pm$0.0197 \\
GPQA Main & 0.3549 $\pm$0.0226 & 0.3147 $\pm$0.0220 & 0.3237 $\pm$0.0221 & 0.3192 $\pm$0.0220 & 0.3125 $\pm$0.0219 & 0.3058 $\pm$0.0218 \\
\hline
\end{tabular}
}
\end{table*}

\section{Vision Models and Multi-Modal Evaluation}
\label{APPENDIX:vision_models}

\begin{table*}[ht]
\centering
\caption{Performance of vision-based models on the TS-Instruct QA benchmark. Metrics are reported as percentages for correct, wrong, and null responses, along with their standard deviations.}
\label{tab:vision_models}
\resizebox{\textwidth}{!}{%
\begin{tabular}{lcccccc}
\toprule
\textbf{Model} & \textbf{Correct Avg (\%)} & \textbf{Correct Std (\%)} & \textbf{Wrong Avg (\%)} & \textbf{Wrong Std (\%)} & \textbf{Null Avg (\%)} & \textbf{Null Std (\%)} \\
\midrule
LLama 3.2-11B-Vision-Instruct & 77.25 & 3.36 & 19.26 & 1.5 & 3.49 & 4.13 \\
Phi 3.5-vision-instruct       & 79.05 & 2.93 & 20.95 & 2.93 & 0.00 & 0.00 \\
\bottomrule
\end{tabular}%
}
\end{table*}

A key contribution of our work is the creation of a multi-modal instruction-tuning and evaluation dataset for time-series reasoning, where the time-series data is derived from \textbf{real-world datasets}. This use of real-world time-series ensures that the generated instructions and tasks are grounded in practical, meaningful data. To achieve this, we rely on vision-based models to link the time-series data to corresponding textual instructions. However, this reliance on vision models comes with the trade-off that their performance on our benchmark is likely \textbf{inflated}, as they are naturally aligned with the dataset generation process. Another important factor is that the multi-modal adapters of these models benefit from the large volumes of multi-modal training data available, which means that the visual adapters are likely more extensively trained in comparison to our multi-modal time-series models.

Recent work by~\citep{merrill2024languagemodelsstrugglezeroshot} provides additional context. It evaluates methods for time-series reasoning on a fully \textbf{synthetic dataset} and demonstrates that vision-based models do not perform significantly better than text-based approaches and, in many cases, are objectively worse. Their results show that on synthetic datasets, vision-based models struggle with time-series-based reasoning. This suggests that vision-based models may not generalize as effectively to other time-series reasoning datasets, particularly those that are synthetically generated.

\paragraph{Summary:}
While our approach benefits from using \textbf{real-world time-series data}, which provides practical grounding and making the dataset relevant to real applications, it comes at the cost of being less suitable for the evaluation of vision-based models for time-series analysis. The alternative approach of using fully synthetic datasets, as in recent work, avoids reliance on visual models but sacrifices the real-world grounding of time-series data. Our methodology highlights this important trade-off.

\textbf{Other Baselines}: 
As discussed in Section 2, our methods and \cite{xie2025chattsaligningtimeseries} are approaching the task from different perspectives and are orthogonal to each other.
While we use the MCQ-2TS dataset introduced by~\cite{merrill2024languagemodelsstrugglezeroshot} in our main experiments, we include extra results here for completeness, as both~\cite{xie2025chattsaligningtimeseries} and~\cite{merrill2024languagemodelsstrugglezeroshot} report evaluations on it. Unfortunately, the experiments in ~\cite{xie2025chattsaligningtimeseries} only utilize a very small, 100-sample subset from the dataset. This data was sampled randomly. In their experiments they report that their 14 billion parameter model achieves roughly $\sim$60\% on this baseline and the base Qwen-14B model achieves $\sim$32\%. We samples 10,000 samples from this dataset and tested out Chat-TS models on this baseline. The base LLama3.1 8B model achieves 36.45\% accuracy and the Chat-TS version achieves 47.56\% which is very reasonable given its size and shows clear performance over the base LLM. Unfortunately our results, the results in \cite{xie2025chattsaligningtimeseries} and the original paper \cite{merrill2024languagemodelsstrugglezeroshot} in which non-GPT models achieve near random performance on this benchmark do not align. Additionally due to the size of the benchmark (well over 100,000 samples) and differences in the sampling strategies (100 random samples vs first 10,000 samples) these results are neither comparable or reproducible. For this reason we do not include them in the main body of our paper and we believe it would be best to generally exclude these benchmarks.

\section{Extended Case Study}
\label{Sec:Extended_Case_Study}
We have selected several examples which show important properties and failure cases of each of the models.
\subsection{TS Instruct QA Gold}
Starting with the QA Benchmark example in Table \ref{TAB: QA example}, we can see that state-of-the-art methods fail in reasoning over a simple trend. The LLama 3.1-8B model identifies a steady increase as opposed to a a significant fluctuation followed by a downward trend. This example highlights the importance of building LLM's for time-series analysis. Existing methods can fail to capture and reason over even basic trends.

Additionally, the models trained without a mix of text only data (TS and PreTS) do not follow the response format explicity specified and therefore it is unlikely the response would be correctly counted towards their score. This is a great example showing that we have instilled knowledge and reasoning abilities over time-series but the instruction following capabilities have deteriorated compared to the models trained on both text and the TS Instruct dataset. Note that this lapse only applies for reasoning on multi-modal data and that the performance on purely text benchmarks has not deteriorated.

\begin{table*}[t]
\centering
\renewcommand{\arraystretch}{1.7}
\resizebox{0.95\textwidth}{!}{
\fontsize{8}{9}\selectfont
\begin{tabular}{m{0.95\textwidth}} 
    \hline
    \parbox{0.95\textwidth}{
        \vspace{2mm}
        \begin{minipage}{0.55\textwidth}
        \vspace{4mm} 
        \textbf{Prompt:} The dataset represents fluctuations in a certain variable over time, likely affected by external factors. Based on this, what could be the most likely interpretation of the trend observed in the time-series data?
        
        A: A steady increase indicates a long-term positive external influence.  
        
        B: Frequent large spikes suggest random noise without any underlying trend. 
        
        C: A systematic drop followed by recovery indicates a seasonal effect.  
        
        D: Significant fluctuations followed by a general downward trend suggest an underlying systemic issue. 
        \vspace{4mm}
        \end{minipage}
        \hfill
        \begin{minipage}{0.4\textwidth}
            \centering
            \includegraphics[width=\textwidth]{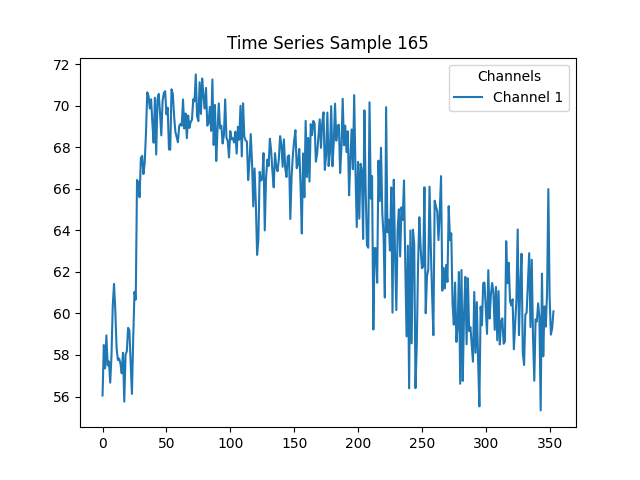}
        \end{minipage}
        \vspace{2mm}
    } \\
    \hline
    \parbox{0.95\textwidth}{
            \vspace{4mm}
            \textbf{Ground Truth Answer:} The correct answer is D: Significant fluctuations followed by a general downward trend suggest an underlying systemic issue. 

            \vspace{2mm}
            \textit{(Optional Explanation):} The observed significant fluctuations along with the overall downward trend indicate potential instability or decline in the underlying variable, pointing to systemic issues rather than simply seasonal effects or random noise.
            \vspace{2mm}
    } \\
    \hline
    
    \parbox{0.95\textwidth}{
        \vspace{4mm}
        \textbf{PreOrcaTS Model Answer:} Based on the time-series data provided, the most likely interpretation of the trend observed is: D: Significant fluctuations followed by a general downward trend suggest an underlying systemic issue.

        \vspace{2mm}
        This interpretation is supported by the presence of multiple peaks and troughs in the data, indicating variability that could be influenced by external factors. The overall downward trend suggests that despite these fluctuations, there is a persistent decline in the variable being measured, which could indicate a systemic issue that needs to be addressed.
        \vspace{4mm}
    } \\
    \hline
    
    \parbox{0.95\textwidth}{
        \vspace{4mm}
        \textbf{OrcaTS Model Answer:} Based on the time-series data provided, the most likely interpretation of the trend observed is: D: Significant fluctuations followed by a general downward trend suggest an underlying systemic issue.

        \vspace{2mm}
        This interpretation is supported by the presence of multiple peaks and troughs, indicating variability in the data, followed by a gradual decline. This pattern could suggest that external factors are influencing the variable, leading to a systemic issue that needs to be addressed.
        \vspace{4mm}
    } \\
    \hline
    
    \parbox{0.95\textwidth}{
        \vspace{4mm}
        \textbf{TS Model Answer:} Based on the analysis of the time-series data, \textcolor{professionalred}{I classify the trend as representing: A steady increase indicates a long-term positive external influence.} 

        \vspace{2mm}
        This suggests that the variable is consistently rising, possibly due to factors like increased demand or improved conditions.
        \vspace{4mm}
    } \\
    \hline

    \parbox{0.95\textwidth}{
        \vspace{4mm}
        \textbf{PreTS Model Answer:} Based on the time-series data, \textcolor{professionalred}{I classify the trend as representing: A steady increase indicates a long-term positive external influence.}

        \vspace{2mm}
        This could be due to consistent growth in demand or an improvement in the underlying process being measured. The gradual rise suggests stability and potential for further growth, which is a positive sign for the variable in question.
        \vspace{4mm}
    } \\
    \hline

    \parbox{0.95\textwidth}{
        \vspace{4mm}
        \textbf{LLama 3.1-8B Model Answer:} A: A steady increase indicates a long-term positive external influence.

        \vspace{2mm}
        \textcolor{professionalred}{The given time-series data shows a general upward trend over time, with some fluctuations. The data points are mostly increasing}, and there are no significant drops or spikes that would indicate a seasonal or random effect. This suggests that there is a long-term positive external influence at play, which is driving the steady increase in the variable being measured. This could be due to various factors such as a growing market, increasing demand, or a positive economic trend.
        \vspace{2mm}
    } \\
    \hline
\end{tabular}
}
\vspace{2mm}
\caption{Example failure case for multi-modal LLMs in interpreting time-series trends. Variability in model interpretations demonstrates potential weaknesses in trend analysis and the risk of hallucinating conclusions from the same dataset.}
\label{TAB: QA example}
\end{table*}

\subsection{TS Instruct}
\label{APPENDIX: TS Instruct Case Study}

Here we show some examples from our TS Instruct benchmark which was generated for the quantitative analysis task. The first example is shown below in Table \ref{tab: Math reasoning Example}. Here we show the a failure case of current time-series analysis methodologies. Because these models rely on normalization as part of the tokenization process, they fail to apply their knowledge to real valued time-series. The models are fed a encoded version of the time-series which after normalization loses its original scale. Even when the model correctly produces the steps used, the final answer is incorrect since it has limited understanding of the original values in the time-series. We also see that the LLama 3.1-8B model tends to create longer and more elaborate answers but that do not necessarily answer the question (for example computing the absolute rate of change instead of the slope).

\begin{table*}[t]
\centering
\renewcommand{\arraystretch}{1.5} 
\small
\begin{tabular}{p{0.95\textwidth}} 
    \hline
    \textbf{Prompt:} Compute the rate of change between the first and last values in the dataset. \\
    \hline
    \begin{minipage}{0.35\textwidth}
        \centering
        \includegraphics[width=\linewidth]{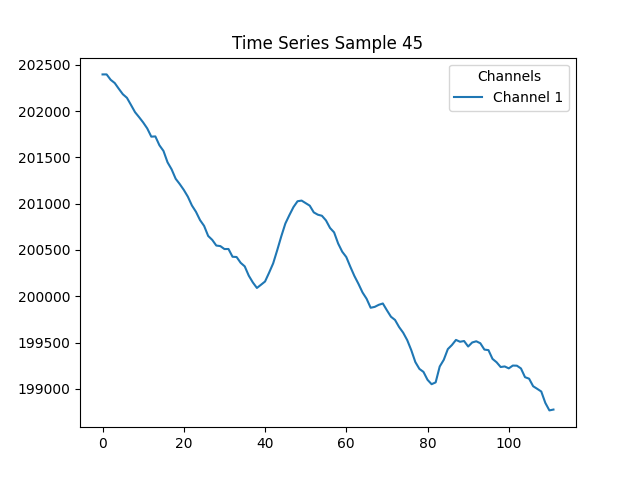}
    \end{minipage}
    \hfill
    \begin{minipage}{0.6\textwidth}
        \textbf{Ground Truth Answer:} Of course! To compute the rate of change, we’ll use the formula:
        \[
        \text{Rate of Change} = \frac{\Delta y}{\Delta x} = \frac{y_{\text{last}} - y_{\text{first}}}{x_{\text{last}} - x_{\text{first}}}
        \]
        From the data, let's assume:
        \begin{itemize}
            \item First value at \(x = 0\) (say 202000)
            \item Last value at \(x = 112\) (say 199000)
        \end{itemize}
        Substituting:
        \[
        \frac{199000 - 202000}{112 - 0} = \frac{-3000}{112} \approx -26.79
        \]
        So, the rate of change is approximately \textbf{-26.79 per time unit}, indicating a decrease over time.
    \end{minipage}
    \\

    \hline
    \parbox{0.95\textwidth}{
        \vspace{2mm}
        \textbf{PreOrcaTS Model Answer:} Sure! To compute the rate of change between the first and last values in the dataset, we can use the formula:
        \( \text{Rate of Change} = \frac{y_{\text{last}} - y_{\text{first}}}{x_{\text{last}} - x_{\text{first}}} \).
        From the time-series data you provided, let's assume the following values:
        \begin{itemize}
            \item \( y_{\text{first}} = 100 \) (value at the first time point)
            \item \( y_{\text{last}} = 200 \) (value at the last time point)
            \item \( x_{\text{first}} = 0 \) (first time point)
            \item \( x_{\text{last}} = 44 \) (last time point)
        \end{itemize}
        Now, substituting these values into the formula:
        \( \text{Rate of Change} = \frac{200 - 100}{44 - 0} = \frac{100}{44} \approx 2.27 \).
        This means that the rate of change in the dataset is approximately \textbf{2.27 units per time point}. This indicates a steady increase in the values over the observed period.
        \vspace{1mm}
    } \\
    \hline
    
    \parbox{0.95\textwidth}{
        \vspace{2mm}
        \textbf{LLama 3.1-8B Model Answer:} To compute the rate of change between the first and last values in the dataset, we need to follow these steps:
        \begin{enumerate}
            \item Identify the first and last values in the dataset.
            \item Calculate the difference between the first and last values.
            \item Divide the difference by the number of time periods (or the number of data points) to get the average rate of change.
        \end{enumerate}
        Let’s identify the first and last values in the dataset:
        \begin{itemize}
            \item The first value is: 1.052
            \item The last value is: 0.2404
        \end{itemize}
        Now, let's calculate the difference between the first and last values:
        \( \text{Difference} = 0.2404 - 1.052 = -0.8116 \).
        Since we are calculating the rate of change, we will use the absolute value of the difference:
        \( |\text{Difference}| = 0.8116 \).
        Now, calculate the average rate of change over 101 data points:
        \( \text{Rate of Change} = \frac{0.8116}{100} = 0.008116 \).
        So, the average rate of change between the first and last values is approximately \textbf{0.008116}. When expressed as a percentage:
        \( \text{Percentage Rate of Change} = 0.008116 \times 100 = 0.81\% \).
        Therefore, the rate of change between the first and last values is approximately \textbf{0.81\% per time period}.
    } \\
    \hline
\end{tabular}
\vspace{2mm}
\caption{Example failure case for multi-modal LLM's in mathematical reasoning. Computing values over the real-life values becomes difficult since the models rely on normalization of the time-series before tokenization. While the process may be generally correct, the models will often make up or hallucinate values causing the actual answers to be inaccurate. }
\label{tab: Math reasoning Example}
\end{table*}

In our next example in Table \ref{reservoir_analysis_example} we show a case study from the decision making tasks in the TS Instruct quantitative analysis. While all of the models correctly identify that the reservoir levels are rising and it may be a good idea to increase the extraction rate with caution, the LLama 3.1-8B SOTA model provides a longer output with several halucinations and incorrect statements. For example it assumes that each data point corresponds to months, however that is not neccissarily true, just that the data spans several months (the x value could be days or weeks). It also provides a good example of the normalization problem wherein the trend under \textbf{Recent increase:} is growing but the values are not correct. It assumes supplementary information like the reservoir level is not critically low and that the increasing reservoir levels mean that the extraction rate is not sufficient (this data was provided by the user and is not directly tied to the water levels).
\begin{table*}[t]
\centering
\renewcommand{\arraystretch}{1.5} 
\resizebox{0.95\textwidth}{!}{
\fontsize{8}{9}\selectfont
\begin{tabular}{p{0.95\textwidth}} 
    \hline
    \parbox{0.95\textwidth}{
        \vspace{2mm}
        \textbf{Prompt:} Hi there! I'm managing a water supply project, and I've been tracking our reservoir levels over the past few months. Currently, I'm debating whether to increase our water extraction rate now to meet growing demand or to maintain our current extraction rate to ensure sustainability. Considering the recent water levels, what do you recommend?
        \vspace{2mm}
    } \\
    \hline

    \parbox{0.95\textwidth}{
        \vspace{2mm}
        \begin{minipage}{0.6\textwidth}
            \textbf{Ground Truth Answer:} After analyzing the reservoir data, it’s evident that the water levels have shown a consistent upward trend over the last several weeks, reaching levels higher than we've seen in recent months. This suggests that there’s ample supply available. However, while the increase looks promising, I recommend that you proceed with caution and slightly increase the extraction rate, rather than a full-scale jump. This way, you can meet the immediate demand without risking long-term sustainability. Balancing rapid extraction with conservation is key here!
            \vspace{2mm}
            \end{minipage}
        \hfill
        \begin{minipage}{0.3\textwidth}
            \centering
            \includegraphics[width=\textwidth]{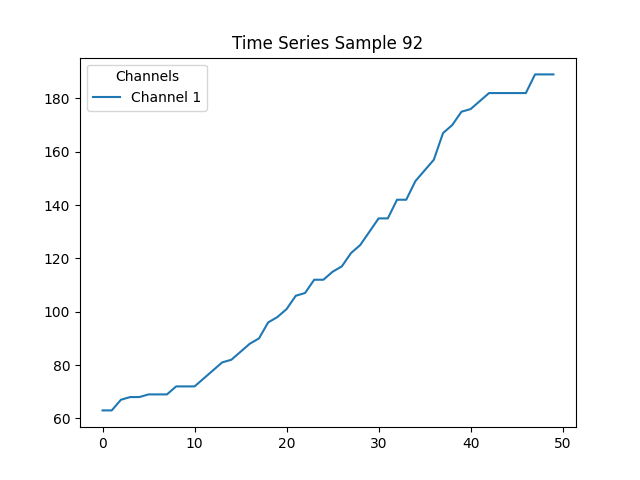}
        \end{minipage}
        \vspace{2mm}
    } \\
    
    \hline

    \parbox{0.95\textwidth}{
        \vspace{2mm}
        \textbf{PreOrcaTS Model Answer:} Based on the time-series data of your reservoir levels, I've observed a significant upward trend over the past few months, indicating a steady increase in water availability. However, there are some fluctuations in the data, suggesting that while the overall trend is positive, there may be periods of volatility. Given the current demand and the rising levels, I recommend increasing your water extraction rate now. This proactive approach will help meet the growing demand while also ensuring that you're making the most of the available resources. However, it's essential to monitor the levels closely to adjust your extraction rate if the trend shifts or if any unexpected changes occur.
        \vspace{2mm}
    } \\
    \hline
    
    \parbox{0.95\textwidth}{
        \vspace{2mm}
        \textbf{OrcaTS Model Answer:} Based on the time-series data you've provided, I've observed a significant upward trend in reservoir levels over the past few months, particularly in the last few weeks. This indicates that the current extraction rate is not causing a depletion of the reservoir, and there's ample water available to meet the growing demand. Therefore, I recommend increasing your water extraction rate now to capitalize on the current surplus and ensure that you can meet the anticipated demand without compromising sustainability in the future.
        \vspace{2mm}
    } \\
    \hline
    
    \parbox{0.95\textwidth}{
        \vspace{2mm}
        \textbf{PreTS Model Answer:} Looking at the time-series data, there's a noticeable upward trend in the reservoir levels over the past few weeks, indicating a recovery from previous lows. However, there are also some fluctuations that suggest potential volatility. Given this context, I recommend increasing the extraction rate now. The rising levels indicate that the reservoir is replenishing, and meeting the growing demand could help manage the supply effectively before any potential downturns occur. It's a strategic move to balance current needs with sustainability.
        \vspace{2mm}
    } \\
    \hline
    
    \parbox{0.95\textwidth}{
        \vspace{2mm}
        \textbf{LLama 3.1-8B Model Answer:} To provide a recommendation, I'll analyze the given time-series data representing the reservoir levels over the past few months. The data is presented as a sequence of numbers, where each number represents the reservoir level at a specific point in time.

        Upon examining the data, I notice the following trends:
        \begin{itemize}
            \item \textbf{Increasing trend}: The reservoir levels have been increasing over time, with some fluctuations. This suggests that the current extraction rate is not depleting the reservoir at an alarming rate.
            \item \textbf{Seasonal fluctuations}: \textcolor{professionalred}{There are noticeable seasonal fluctuations in the reservoir levels, with higher levels during the summer months (e.g., July, August) and lower levels during the winter months (e.g., January, February).}
            \item \textbf{Recent increase}: \secondplace{The reservoir levels have been increasing rapidly over the past few months, with a significant jump in the last few data points (e.g., 9.7, 9.7, 9.7, 9.7, 9.7, 10.19, 10.19, 10.19).}
        \end{itemize}

        Considering the increasing trend and recent increase in reservoir levels, I recommend \textbf{increasing the water extraction rate} to meet the growing demand. This decision is based on the following reasoning:
        \begin{itemize}
            \item \textbf{Current sustainability}: \textcolor{professionalred}{The reservoir levels are not critically low}, and the current extraction rate is not depleting the reservoir at an alarming rate.
            \item \textbf{Growing demand}: \textcolor{professionalred}{The increasing trend in reservoir levels suggests that the current extraction rate is not sufficient to meet the growing demand.}
            \item \textbf{Opportunistic increase}: With the recent increase in reservoir levels, it is an opportune time to increase the extraction rate to meet the growing demand without compromising the sustainability of the reservoir.
        \end{itemize}

        To ensure the sustainability of the reservoir, it is essential to monitor the reservoir levels closely and adjust the extraction rate as needed. This may involve:
        \begin{itemize}
            \item Regular monitoring of reservoir levels to ensure they remain within a sustainable range.
            \item Adjusting extraction rates based on the reservoir levels, accounting for growing demand and seasonal fluctuations.
            \item Implementing conservation measures, such as reducing non-essential water usage or implementing water-saving technologies, to reduce the demand on the reservoir.
        \end{itemize}
        \vspace{2mm}
    } \\
    \hline
\end{tabular}
}
\vspace{2mm}

\caption{Example of model outputs for reservoir level analysis. Models provide varying recommendations based on the same time-series data, highlighting their reasoning and suggestions for water extraction rate management.}
\label{reservoir_analysis_example}
\end{table*}

\section{TS-Instruct QA Gold Dataset Evaluation}
\label{App:mturk}
A first pass of the evaluation dataset was during using Amazon Mechanical-Turk\cite{mturk}. Due to the high volume of questions in the original generated dataset we did not do inter-worker evaluations, however each worker was required to have the following two criteria: $\geq$ 90 \% approval, $\geq$ 100 HITs. Each item received one annotation and we manually audited the reviews to ensure each selected sample that was kept for our evaluation dataset was of high quality. Compensation was $\$0.02$ USD per sample.

Each evaluation was shown an image of the time-series along with the question, the options and the intended correct answer option. Based on this information they selected:
\begin{itemize}
    \item Does the correct answer most accuractely describe the time-series?
    \item Is the correct answer the ONLY correct answer?
    \item The domain of the time-series
    \item Clarity of the question
    \item (Optional) Notes about the question
\end{itemize}

\end{document}